\documentclass{article} 
\usepackage{iclr2027_conference,times}

\usepackage{amsmath,amsfonts,bm}

\def\eqref#1{equation~\ref{#1}}

\def\1{\bm{1}}

\DeclareMathAlphabet{\mathsfit}{\encodingdefault}{\sfdefault}{m}{sl}
\SetMathAlphabet{\mathsfit}{bold}{\encodingdefault}{\sfdefault}{bx}{n}

\usepackage{hyperref}
\usepackage{url}
\usepackage{bbm}
\usepackage{amsmath}
\usepackage{graphicx}

\usepackage{booktabs}
\usepackage{multirow}
\usepackage[table]{xcolor}

\usepackage{titletoc} 
\usepackage{subcaption} 
\usepackage{float} 

\usepackage{enumitem}

\usepackage{caption}

\usepackage{algorithm}
\usepackage{algpseudocode}

\usepackage[most]{tcolorbox}

\definecolor{promptpink}{HTML}{D97D96}
\definecolor{promptbg}{HTML}{FFF7F9}
\definecolor{promptblue}{HTML}{1769D2}

\newcommand{\pvar}[1]{\textcolor{promptblue}{\{#1\}}}

\newtcolorbox{prompttemplate}[1]{
  enhanced,
  breakable,
  colback=promptbg,
  colframe=promptpink,
  boxrule=0.8pt,
  arc=2.5mm,
  left=4mm,
  right=4mm,
  top=5mm,
  bottom=3mm,
  title={Prompt Template --- #1},
  fonttitle=\bfseries\large\sffamily,
  coltitle=white,
  boxed title style={
    colback=promptpink,
    colframe=promptpink,
    boxrule=0pt,
    arc=0pt
  },
  attach boxed title to top left={
    xshift=5mm,
    yshift=-2.5mm
  },
  before skip=8pt,
  after skip=10pt
}

\newcommand{\teacherfeedback}{%
\tcblower
\normalfont\small\itshape
Teacher feedback context:\par\smallskip
\ttfamily\upshape
You are the feedback-conditioned teacher in on-policy self-distillation.
The earlier student attempt will appear as the assistant continuation.
Use the structured feedback context to revise the continuation distribution
for the same task. Preserve useful reasoning tokens and end with the task's
normal final-answer line. Do not quote or describe the context.\par\medskip

Task:\par
\pvar{dataset task prompt}\par\medskip

Structured feedback context:\par
[Context block 1]\par
\pvar{verifier result}\par
[Context block 2]\par
\pvar{parser record}\par
[Context block 3]\par
\pvar{verifier provenance}\par
[Context block 4]\par
\pvar{response-format diagnostics}%
}

\title{Fisher-Informed Recalibration for Feedback-Based On-Policy Self-Distillation of LLMs}

\author{Seohyun Lee\textsuperscript{\rm 1},  
Dong-Jun Han\textsuperscript{\rm 2},
\textbf{Seyyedali Hosseinalipour}\textsuperscript{\rm 3}, 
\textbf{Christopher G. Brinton}\textsuperscript{\rm 1} \\
\textsuperscript{\rm 1}Purdue University, 
\textsuperscript{\rm 2}Yonsei University, 
\textsuperscript{\rm 3}University at Buffalo-SUNY\\
\texttt{lee3296@purdue.edu}, \texttt{djh@yonsei.ac.kr}, \\
\texttt{alipour@buffalo.edu}, \texttt{cgb@purdue.edu}\\
}

\iclrfinalcopy 
\begin{document}

\maketitle

\begin{abstract}
Feedback-based on-policy self-distillation has emerged as a promising approach for enabling foundation models, more specifically Large Language Models (LLMs), to learn from their own outputs under external feedback, with a single model serving as both teacher and student. However, such methods can exhibit unstable optimization, conducive to performance collapse during training. To address this limitation, we propose \textbf{FIRE} (\textbf{F}isher-\textbf{I}nformed \textbf{RE}calibration), \textit{a dual-branch framework that recalibrates the supervision applied to correct and incorrect on-policy outputs} during fine-tuning. For correct responses, FIRE replaces self-distillation with re-weighted on-policy SFT, while for incorrect ones FIRE identifies feedback components that disproportionately influence the teacher-induced update and recalibrates the feedback-conditioned target accordingly. Both branches are influenced by a token-level radius derived in part from a softmax Fisher trace. FIRE separates \emph{which direction} feedback should move the model from \emph{how far} the model should move in that direction, while leaving well-behaved feedback supervision unchanged. Our experiments demonstrate that FIRE provides substantially more stable self-distillation while maintaining strong downstream performance, particularly in settings where standard feedback-conditioned distillation becomes unstable.
\end{abstract}

\section{Introduction}
Foundation Models \citep{bommasani2021opportunities}, which are models trained on large amounts of data, have become increasingly influential and adopted in practice, with Large Language Models (LLMs) particularly enjoying widespread adoption \citep{minaee2024large}. In practice, many available LLMs are pre-trained \citep{grattafiori2024llama, yang2025qwen3}, with fine-tuning being a method for adapting these models to specific use cases, updating only a small number of parameters relative to the total model \citep{hu2021lora, zhang2025parameter}.

In parallel, fine-tuning such LLMs from \textit{their own outputs} has become an emerging research direction \citep{lee2026self, hubotter2026reinforcement}. In particular, feedback-based on-policy self-distillation \citep{zhao2026self} has emerged as an increasingly popular paradigm. The model first generates an initial response, receives feedback (e.g., user interactions), and conditions on this feedback to produce a revised response. The feedback-conditioned model then serves as a teacher, with the original model optimized to minimize the Kullback-Leibler (KL) \citep{kullback1951information} divergence between its output distribution and the feedback-conditioned distribution.

\paragraph{Challenges.}
Despite its promise, feedback-based on-policy self-distillation fine-tuning presents multiple challenges. Firstly, not every on-policy response should be trained from feedback-conditioned teacher supervision in the same manner. Applying self-distillation to responses that are already correct can be unnecessary and may contribute to model divergence~\citep{zheng2026scope}, since the response itself already provides an unambiguous positive learning signal. Conversely, when a response is incorrect, feedback is necessary to determine how the model should revise its prediction, but directly matching the feedback-conditioned teacher blindly, whose revised outputs may still be incorrect, can potentially introduce excessively large and poorly directed updates, which also induce performance collapse \citep{lee2026self}. Moreover, standard KL-based self-distillation treats the entire feedback context as an infallible source of supervision and provides no mechanism for determining whether a particular feedback component disproportionately drives the resulting update. Lastly, teacher-student discrepancy could substantially affect optimization depending on the student's local uncertainty and the learning rate. These observations motivate two complementary requirements: the training procedure should distinguish between correct and incorrect on-policy trajectories, and it should constrain feedback-induced updates according to the student's local predictive geometry while preserving informative feedback whenever its induced update is already well behaved.

\subsection{Contributions}

To address these challenges, we propose \textbf{FIRE} 
(\textbf{F}isher-\textbf{I}nformed \textbf{RE}calibration), a \textit{dual-branch framework for stabilizing feedback-based on-policy self-distillation fine-tuning of LLMs}. FIRE distinguishes between correct and incorrect on-policy responses and recalibrates their training 
signals according to the student's local predictive geometry. For correct responses, FIRE replaces feedback-conditioned self-distillation with Fisher-informed re-weighted on-policy supervised fine-tuning (SFT). For incorrect responses, FIRE attributes the influence 
of individual feedback components and recalibrates the resulting teacher target before distillation. Overall, our main contributions are as follows:

\begin{itemize}[leftmargin=*, itemsep=2pt, topsep=2pt, parsep=0pt]
    \item \textit{Dual-branch on-policy self-distillation.}
    We introduce an outcome-routed framework that applies differing supervision to correct and incorrect on-policy responses. In contrast with existing work that adopts correct-branched SFT, correct responses are optimized with gradient and Fisher-informed re-weighted on-policy SFT. Incorrect responses retain feedback-conditioned self-distillation, allowing feedback to guide revision only where it is needed.

    \item \textit{Fisher-informed feedback recalibration.}
    For incorrect responses, FIRE uses leave-one-feedback-field-out counterfactuals to quantify how individual feedback components influence the teacher-induced update in the gradient space. We combine this attribution with a token-level radius derived from the student's softmax Fisher trace and the learning rate, separating \emph{which direction} feedback should move the model from \emph{how far} the model should move in that direction. Importantly, FIRE leaves the original 
    full-feedback supervision unchanged whenever its induced update already lies within the permitted radius.

    \item \textit{Minimal hyperparameter tuning.} Outside of the \textit{nominal rate}, which controls how aggressively the radius contracts as the learning rate increases, FIRE introduces no additional hyperparameters. Therefore, FIRE does not require meticulous hyperparameter tuning to fully take advantage of its performance improvements.

    \item \textit{Bounded and stable self-distillation.}
    We show that FIRE bounds the token-level output-logit gradient in both the correct and incorrect branches by the same student-dependent radius. Empirically, we additionally show that this recalibration substantially improves training stability over standard feedback-conditioned self-distillation and numerous relevant baselines, particularly when standard self-distillation becomes unstable.
\end{itemize}

\section{Related Work}

\paragraph{Knowledge Distillation (KD).} Knowledge Distillation (KD) \citep{hinton2015distilling} is a common technique employed in deep learning that is designed to distill capabilities from one model (the teacher) to another (the student) \citep{lee2025tap}. More specifically, the student seeks to match the teacher's output distribution through a KL-divergence \citep{kullback1951information} based loss. KD-based training has been demonstrated to be an effective method of aligning the student's distribution to the teacher's \citep{phuong2019towards}, and has been used extensively in foundation model settings such as LLMs \citep{xu2024survey}. Therefore, such distillation has become an increasingly adopted option for models trained on synthetic outputs generated by themselves or another model \citep{jang2026stable, xie2026trust, xing2026trust}. In our work, we consider the first option. 

\paragraph{On-Policy Self-Distillation and Stabilization.} On-Policy Self-Distillation is a KD-based methodology in which a model learns from supervision constructed around \textit{its own generated outputs} \citep{zhao2026self}. This is in contrast with on-policy distillation, which assumes the teacher and student can be differing models \citep{xie2026trust, xing2026trust}. For on-policy self-distillation, the current student policy first generates an output, producing an on-policy trajectory \citep{shenfeld2026self}. Then this original generation and feedback context is provided again, which produces a new target output conditioned on the student's behavior and feedback \citep{lee2026self}, which is what is distilled upon. As the student changes throughout training, both the trajectories presented to the teacher and the resulting distillation targets likewise change. With existing research demonstrating distilling on all samples can induce performance collapse~\citep{li2026unifying, zheng2026scope}, existing literature seeks to stabilize the self-distillation process by dual-branching correct and incorrect trajectories, with \citet{li2026unifying} and \citet{zheng2026scope} considering Group Relative Policy Optimization (GRPO)~\citep{shao2024deepseekmath} and a perplexity-based version of supervised fine-tuning (SFT)~\citep{ouyang2022training} on correct trajectories respectively. In addition, works such as \citet{li2026demopsd} measure the disagreement between the teacher and student, geometrically shifting the teacher closer to the student when the differences are unreconcilable. Outside of the self-distillation setting, stabilization of on-policy distillation has also been explored under the idea of a trust region, constraining unreliable and overly large updates~\citep{xie2026trust, xing2026trust}. Compared to prior work, FIRE applies Fisher-informed re-weighted on-policy SFT to correct trajectories, while recalibrating incorrect-trajectory self-distillation through feedback-field attribution and a Fisher-informed update radius. Further details are provided in Sec.~\ref{sec:method} and Appendix~\ref{appendix:novelty_info}.

\paragraph{Parameter Efficient Fine-tuning (PEFT).}
When fine-tuning foundation models, it is desirable to limit the number of parameters which are updated for computational efficiency. For this, parameter efficient fine-tuning methods (PEFT) are particularly popular and have demonstrated good performance across a wide range of domains \citep{fang2025federated, liu2024moe}. In addition, it is demonstrated to be useful in resource-constrained settings \citep{fang2025federated, lee2025tap}. Common methods to employ PEFT include prefix-tuning \citep{li2021prefix} and LoRA \citep{hu2021lora}, with the latter becoming the most popular option and the approach we adopt. 

\section{Proposed Methodology}
\label{sec:method}

\subsection{Feedback-Based On-Policy Self-Distillation}
\label{sec:sdpo_setup}

We define $\pi_{\boldsymbol{\theta}_k}$ as the trainable \emph{student} at step $k$, and denote $\pi_{\bar{\boldsymbol{\theta}}_k}$ as an exponential moving average (EMA) \emph{teacher}, which we adopt from~\citet{hubotter2026reinforcement} for stability. The teacher is initialized from the student, $\bar{\boldsymbol{\theta}}_0=\boldsymbol{\theta}_0$, and is updated only through EMA rather than backpropagation. Given a prompt $\mathbf{x}\sim\mathcal{D}$, with $\mathcal{D}$ denoting an incoming data stream (e.g., user interaction), the current student first samples one on-policy response, i.e., 
\begin{equation}
\mathbf{y}=(y_1,\ldots,y_{T})
\sim
\pi_{\boldsymbol{\theta}_k}(\cdot\mid\mathbf{x}).
\label{eq:sdpo_rollout}
\end{equation}
A feedback mechanism then produces privileged feedback
$\mathbf{c}=\mathcal{F}(\mathbf{x},\mathbf{y})$ about the sampled response. For example, $\mathbf{c}$ may contain a critique, correction, reasoning hint, or reference information. This feedback is available only during training and is provided to the teacher, not the student. Then, for response position $t$, both models score the same sampled token prefix $\mathbf{y}_{<t}$. We define the live student distribution $p_{\boldsymbol{\theta}}^t(v)$ and detached step-$k$ teacher distribution $q_{\mathbf{c},k}^t(v)$ as
\begin{equation}
\begin{aligned}
p_{\boldsymbol{\theta}}^t(v)
&=
\pi_{\boldsymbol{\theta}}
(v\mid\mathbf{x},\mathbf{y}_{<t}),\ &
q_{\mathbf{c},k}^t(v)
=
\operatorname{sg}\left[
\pi_{\bar{\boldsymbol{\theta}}_k}
(v\mid\mathbf{x},\mathbf{c},\mathbf{y}_{<t})
\right],
\qquad v\in\mathcal{V},
\end{aligned}
\label{eq:sdpo_distributions}
\end{equation}
where $\mathcal{V}$ is the vocabulary and $\operatorname{sg}$ denotes a stop-gradient. 
At step $k$, $p_{\boldsymbol{\theta}}^t(v)$ is instantiated as $\boldsymbol{\theta} = \boldsymbol{\theta}_k$ to denote that it is the live distribution, while $q_{\mathbf{c},k}^t(v)$ is a detached target. We omit $v$ when referring to either distribution.  
Importantly, evaluating $q^t_{\mathbf{c},k}$ does not require an additional autoregressive rollout. Because the complete student trajectory $\mathbf{y}$ is already available, $q^t_{\mathbf{c},k}$ is obtained simultaneously through a single teacher forward pass under the causal mask. Therefore, the EMA teacher scores the existing student trajectory rather than generating a new response.
Next, the student is trained through reverse-KL distillation, i.e.,
\begin{equation}
\mathcal{L}_{\mathrm{KL}}(\boldsymbol{\theta})
=
\frac{1}{T}
\sum_{t=1}^{T}
\mathrm{KL}\!\left(
p_{\boldsymbol{\theta}}^t
\,\middle\|\,
q_{\mathbf{c},k}^t
\right).
\label{eq:sdpo_objective}
\end{equation}

After backpropagating, the student is updated first. Then the EMA teacher tracks the student, i.e., 
\begin{equation}
    \boldsymbol{\theta}_{k+1}=
\operatorname{OptStep}\left(
\boldsymbol{\theta}_k,
\nabla_{\boldsymbol{\theta}_k}
\mathcal{L}_{\mathrm{KL}}
\right),
\qquad
\bar{\boldsymbol{\theta}}_{k+1}
=
\mu\bar{\boldsymbol{\theta}}_k
+(1-\mu)\boldsymbol{\theta}_{k+1},
\label{eq:sdpo_updates}
\end{equation}
where $\operatorname{OptStep}$ is the update rule (e.g., SGD with momentum) and $\mu\in[0,1)$ is the EMA decay. 

\subsection{FIRE: \textbf{F}isher-\textbf{I}nformed \textbf{Re}calibration}
\label{sec:fire}

\begin{figure*}[t]
    \centering
    \includegraphics[width=1.0\linewidth]{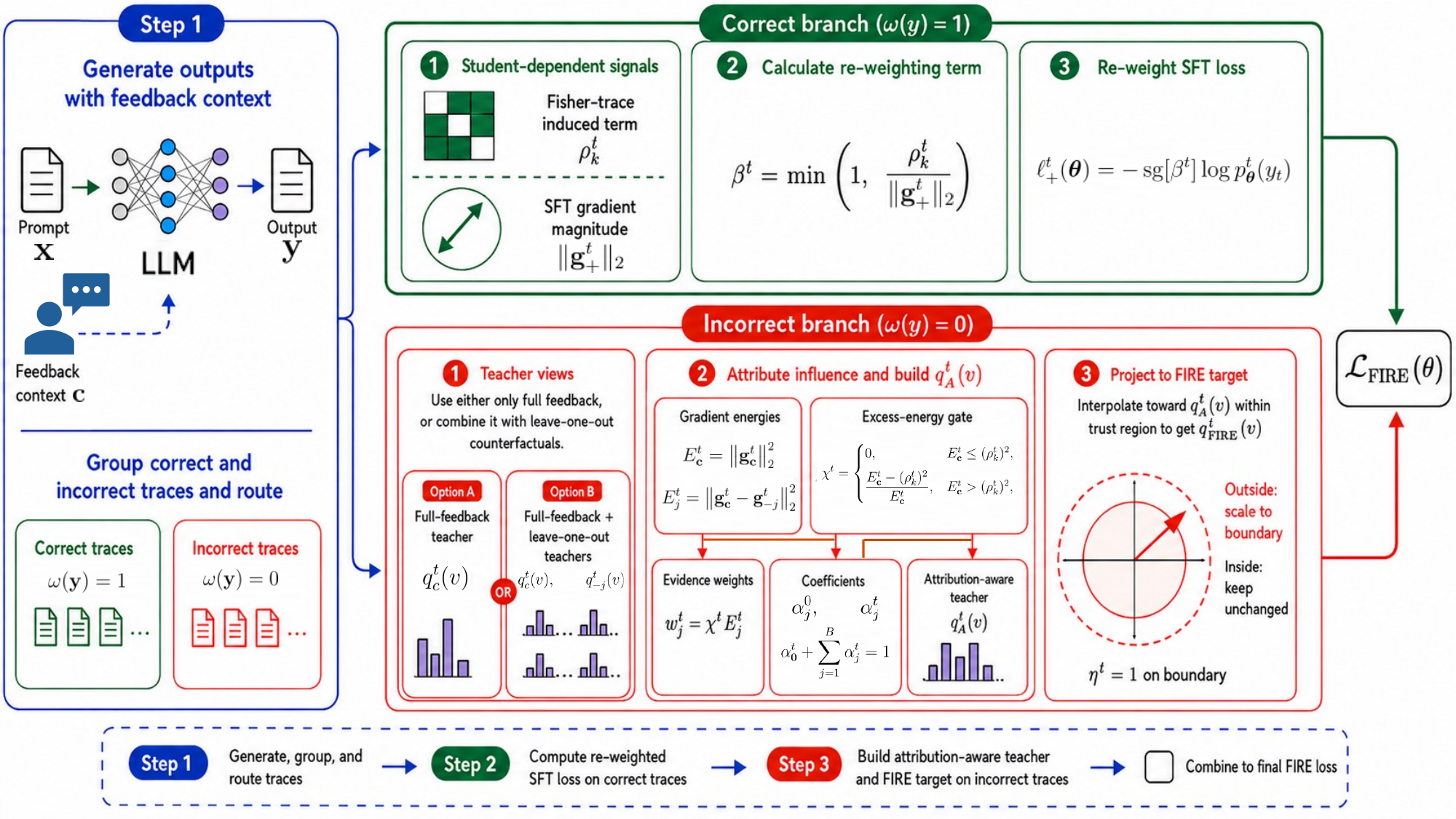}
    \caption{The fine-tuning process of the proposed FIRE methodology. Firstly, we separate traces produced from the model into correct and incorrect groups. Next, Fisher-informed re-weighted on-policy SFT is conducted on correct traces while incorrect traces engage in recalibrated self-distillation, re-weighting based on leave-one-out counterfactuals, full feedback, and gradient insights. Both branches are influenced by the Fisher-informed term $\rho_{k}^t$. The two branches are combined to form the final loss. Specifics on differences from similar works is found in Appendix \ref{appendix:novelty_info}.}
    \label{fig:FIRE_method}
\end{figure*}

In this subsequent section, we outline the training procedure of FIRE, which modifies how the feedback-conditioned supervision in Sec.~\ref{sec:sdpo_setup} is applied. FIRE routes correct and incorrect rollouts differently while constraining both branches with a shared student-dependent output-gradient radius. Complete derivations are provided in Appendix~\ref{app:fire_details}.

\paragraph{Step 1: Outcome routing and teacher views.}
For responses that are already correct, existing work shows that applying self-distillation to those samples can lead to collapse~\citep{li2026unifying, zheng2026scope}. Therefore, after sampling $\mathbf{y}$, a verifier $\omega(\mathbf{y})\in\{0,1\}$ indicates whether the initial output is correct. Correct responses receive Fisher-informed re-weighted on-policy SFT, whereas incorrect responses receive feedback-conditioned distillation. Intuitively, a verified response already provides an unambiguous positive target, while an incorrect response requires the teacher's feedback to determine how the prediction should change.

For target construction, we define $p^t(v)=\operatorname{sg}[p_{\boldsymbol{\theta}_k}^t(v)]$ as the detached current student distribution. On an incorrect response, FIRE represents the feedback as fields $\mathbf{c}=(b_1,\ldots,b_B)$, where $B$ is the number of feedback fields. For notational simplicity, we suppress step index $k$ on detached targets, writing $q_{\mathbf{c}}^t(v):=q_{\mathbf{c},k}^t(v)$. We use $p^t$ and $q_{\mathbf{c}}^t$ without an argument to denote the corresponding distributions.
The EMA teacher produces this full-feedback target from \eqref{eq:sdpo_distributions}, together with the leave-one-field-out targets
\begin{equation}
q_{-j}^t(v)
=
\operatorname{sg}\left[
\pi_{\bar{\boldsymbol{\theta}}_k}
(v\mid\mathbf{x},\mathbf{c}_{\setminus j},\mathbf{y}_{<t})
\right],
\label{eq:loo_teacher}
\end{equation}
where $\mathbf{c}_{\setminus j}$ removes feedback field $b_j$. Each leave-one-field-out view is a counterfactual that asks \textit{how the teacher's supervision would change without field} $b_j$, allowing FIRE to attribute influence to individual feedback components rather than treating the context as one block. We demonstrate that despite this overhead, our method achieves the best time-accuracy tradeoff in Appendix \ref{appendix_subsec:timevsacc}. For any detached target $q^t$, we define its induced student-logit gradient as
\begin{equation}
\mathbf{g}^t(q^t)
:=
\left.
\nabla_{\mathbf{z}^t}
\mathrm{KL}\!\left(
\operatorname{softmax}(\mathbf{z}^t)
\,\middle\|\,
q^t
\right)
\right|_{\mathbf{z}^t=\mathbf{z}_k^t}.
\label{eq:induced_gradient}
\end{equation}
where $\mathbf{z}_k^t$ denotes the current student logits at token position $t$. This converts each teacher distribution into the actual local training direction it would impose on the student. FIRE performs attribution in the gradient space since a change in teacher probabilities matters only in that it changes the update the student will receive. FIRE bounds this local signal using
\begin{equation}
\rho_k^t
=
\frac{
\sqrt{1-\lVert p^t\rVert_2^2}
}{
\max(1,a_k/a_0)
},
\label{eq:fire_radius}
\end{equation}
where $a_k$ is the current learning rate and $a_0$ is a hyperparameter we call the \textit{nominal rate}. The numerator is the square root of the softmax Fisher trace, which measures how spread out the student’s prediction is and gives FIRE a natural reference for choosing the update size. When the student places most of its probability on only a few tokens, the radius is smaller. When its probability is spread across many tokens, the radius is larger. The denominator further reduces the radius when the learning rate is higher, preventing the larger learning rate from making the update too strong and causing divergence.
Importantly, while existing work also considers trust-region-style stabilization for on-policy distillation~\citep{xie2026trust,xing2026trust}, these methods consider \textit{conventional teacher--student distillation rather than self-distillation} and \textit{do not use the student's softmax Fisher trace to calibrate the local update radius} (additional details in Appendix \ref{appendix:novelty_info}).
 The explicit derivation is provided in Appendix~\ref{app:fisher_derivation}.

\paragraph{Step 2: Correct branch training.}
Next, for $\omega(\mathbf{y})=1$, the standard on-policy SFT gradient is $\mathbf{g}_+^t=p^t-\mathbf{e}_{y_t}$, where $\mathbf{e}_{y_t}$ is the one-hot vector for token $y_t$. FIRE then re-weights the SFT loss via
\begin{equation}
\beta^t
=
\min\!\left(
1,
\frac{\rho_k^t}{\lVert\mathbf{g}_+^t\rVert_2}
\right),
\qquad
\ell_+^t(\boldsymbol{\theta})
=
-\operatorname{sg}[\beta^t]
\log p_{\boldsymbol{\theta}}^t(y_t).
\label{eq:fire_correct}
\end{equation}
Then, with \eqref{eq:fire_correct}, we balance the uncertainty of the update ($\rho_k^t$) with the magnitude of the gradient ($\lVert\mathbf{g}_+^t\rVert_2$). Therefore, FIRE on correct traces balances student uncertainty with the magnitude of the induced gradient, unlike existing work~\citep{zheng2026scope}, which weights correct trajectories using student perplexity, in contrast with \textit{an explicit Fisher-informed gradient-magnitude constraint} (additional details in Appendix \ref{appendix:novelty_info}).

\paragraph{Step 3: Incorrect branch training.}
For incorrect responses, i.e., $\omega(\mathbf{y})=0$, we let $\mathbf{g}_{\mathbf{c}}^t=\mathbf{g}^t(q_{\mathbf{c}}^t)$ and $\mathbf{g}_{-j}^t=\mathbf{g}^t(q_{-j}^t)$. For an incorrect trace, we define the full-teacher gradient energy and the leave-one-field-out influence energy as
\begin{equation}
E_{\mathbf{c}}^t
=
\left\lVert
\mathbf{g}_{\mathbf{c}}^t
\right\rVert_2^2,
\qquad
E_j^t
=
\left\lVert
\mathbf{g}_{\mathbf{c}}^t
-
\mathbf{g}_{-j}^t
\right\rVert_2^2,
\label{eq:fire_gradient_energies}
\end{equation}
with $E_{\mathbf{c}}^t$ measuring the total strength of the update requested by the complete feedback-conditioned teacher, while $E_j^t$ measures how much removing field $b_j$ changes that update. A large $E_j^t$ indicates influence, but does not by itself imply that the field is harmful. We require that the full teacher must first violate the local budget before FIRE attenuates any field. FIRE therefore defines the excess-energy gate
\begin{equation}
\chi^t
=
\begin{cases}
0,
&
E_{\mathbf{c}}^t
\leq
(\rho_k^t)^2,
\\[6pt]
\displaystyle
\frac{
E_{\mathbf{c}}^t-(\rho_k^t)^2
}{
E_{\mathbf{c}}^t
},
&
E_{\mathbf{c}}^t
>
(\rho_k^t)^2,
\end{cases}
\qquad
w_j^t
=
\chi^t E_j^t.
\label{eq:fire_evidence}
\end{equation}
The gate separates influence from unsafe influence. If the full teacher already lies within the permitted energy radius, then $\chi^t=0$ and the full-feedback target is retained exactly. If not, $\chi^t\in(0,1)$ measures the fraction of its gradient energy lying beyond the boundary. Multiplying each $E_j^t$ by this common fraction preserves the relative ranking of field influence while making the overall attenuation strength proportional to the severity of the budget violation. After computing the field weights $\{w_j^t\}_{j=1}^{B}$, FIRE normalizes the full-teacher anchor and leave-one-field-out evidence, i.e.,
\begin{equation}
Z^t
=
(\rho_k^t)^2
+
\sum_{j=1}^{B}w_j^t,
\qquad
    \alpha_0^t
=
\frac{(\rho_k^t)^2}{Z^t},
\qquad
\alpha_j^t
=
\frac{w_j^t}{Z^t}.
\label{eq:fire_attribution_weights}
\end{equation}
The coefficients are nonnegative and satisfy $\alpha_0^t+\sum_{j=1}^{B}\alpha_j^t=1$. The term $(\rho_k^t)^2$ acts as evidence for retaining the complete teacher, while each $w_j^t$ acts as evidence for moving toward the counterfactual teacher in which field $b_j$ is absent. Because both quantities are measured in gradient-energy units, their normalization yields dimensionless weights that balance retaining the full-feedback teacher against reducing reliance on influential fields. In the degenerate case $Z^t=0$, we adopt the convention $\alpha_0^t=1$ and $\alpha_j^t=0$. With these coefficients, FIRE constructs an attribution-aware teacher
\begin{equation}
q_A^t(v)
=
\frac{
q_{\mathbf{c}}^t(v)^{\alpha_0^t}
\prod_{j=1}^{B}
q_{-j}^t(v)^{\alpha_j^t}
}{
\sum_{u\in\mathcal{V}}
q_{\mathbf{c}}^t(u)^{\alpha_0^t}
\prod_{j=1}^{B}
q_{-j}^t(u)^{\alpha_j^t}
},
\label{eq:fire_attribution_target}
\end{equation}
with the geometric mean utilized because averaging in log-probability space induces the corresponding weighted combination of reverse-KL logit gradients, so the attribution weights have a direct interpretation in the resulting optimization direction (see Appendix~\ref{app:field_attribution}).

Then we measure the gradient induced by the attribution-aware target and define the radial contraction coefficient
\begin{equation}
\eta^t
=
\begin{cases}
1,
&
\left\lVert
\mathbf{g}^t(q_A^t)
\right\rVert_2
\leq
\rho_k^t,
\\[6pt]
\displaystyle
\frac{
\rho_k^t
}{
\left\lVert
\mathbf{g}^t(q_A^t)
\right\rVert_2
},
&
\left\lVert
\mathbf{g}^t(q_A^t)
\right\rVert_2
>
\rho_k^t,
\end{cases}
\label{eq:fire_contraction_coefficient}
\end{equation}
where $\eta^t$ is the interpolation coefficient that keeps the attribution-corrected gradient within the local radius. It equals one when no contraction is necessary and decreases only enough to place an oversized signal on the boundary. Finally, FIRE interpolates geometrically between the detached student and the attribution-aware teacher:
\begin{equation}
q_{\mathrm{FIRE}}^t(v)
=
\frac{
p^t(v)^{1-\eta^t}
q_A^t(v)^{\eta^t}
}{
\displaystyle\sum_{u\in\mathcal{V}}
p^t(u)^{1-\eta^t}
q_A^t(u)^{\eta^t}
}.
\label{eq:fire_final_target}
\end{equation}
This construction scales the induced gradient exactly as $\mathbf{g}^t(q_{\mathrm{FIRE}}^t)=\eta^t\mathbf{g}^t(q_A^t)$. Field attribution therefore determines which feedback-dependent direction should be followed, while the final interpolation determines how far the student may move in that direction. Computing $\eta^t$ after attribution guarantees that the final signal satisfies the radius. The corresponding derivation is provided in Appendix~\ref{app:failure_projection}. Importantly, FIRE's incorrect branch training differs from recent work that considers teacher-student geometric interpolation~\citep{li2026demopsd} by considering (i) \textit{how each individual feedback field influences the updates} and (ii) \textit{readjusting for excess gradient energy}. Additional specifics are provided in Appendix \ref{appendix:novelty_info}.

\paragraph{Step 4: FIRE Loss Function.} As a last step, we define $m_{i,t}$ to mark valid response tokens and $M_i=\sum_t m_{i,t}$, with $i$ denoting the batch sample index. For a minibatch of $N$ responses, the final objective of FIRE for the student update is
\begin{equation}
\begin{aligned}
\mathcal{L}_{\mathrm{FIRE}}(\boldsymbol{\theta})
=
\frac{1}{N}\sum_{i=1}^{N}\frac{1}{M_i}\sum_t m_{i,t}
\Big[
&\omega(\mathbf{y}_i)\ell_{+,i}^t(\boldsymbol{\theta})
+
\bigl(1-\omega(\mathbf{y}_i)\bigr)
\mathrm{KL}\!\left(
p_{\boldsymbol{\theta},i}^t
\,\middle\|\,
\operatorname{sg}[q_{\mathrm{FIRE},i}^t]
\right)
\Big],
\end{aligned}
\label{eq:fire_objective}
\end{equation}
with both branch gradients bounded by $\rho_{k,i}^t$, as shown in Appendix~\ref{app:unified_bound}. The EMA teacher update remains unchanged from \eqref{eq:sdpo_updates}.

\section{Experimental Results}\label{sec:experiments}

\paragraph{Datasets, Models, Basic Setup.} For evaluation of the FIRE methodology, we consider Qwen2.5-Instruct from the well-established Qwen family of foundation models~\citep{qwen2.5}. Specifically, we consider sizes of 1.5B and 3B, with a larger model size (Qwen2.5-14B) considered in Appendix \ref{appendix_sec:14b}. For Qwen2.5-1.5B-Instruct, we consider the GSM8K \citep{cobbe2021gsm8k}, AQuA-RAT \citep{ling2017program}, and ASDiv \citep{miao-etal-2020-diverse} datasets. For Qwen2.5-3B-Instruct, we consider ARC-Challenge \citep{allenai:arc}, SuperGLUE WiC \citep{wang2019superglue, pilehvar2018wic}, and WinoGrande \citep{sakaguchi2021winogrande}. We consider disjoint datasets across model sizes because some benchmarks become saturated at larger scales, exhibiting little to no improvement over the pretrained model during fine-tuning. For all experiments, we conduct linear warmup on the learning rate, with cosine decay after the peak learning rate is reached. Additionally, all experiments utilize AdamW as the optimizer \citep{loshchilov2017decoupled}. We utilize LoRA \citep{hu2021lora} as the PEFT method. For time efficiency, we evaluate on subsets of the test data. These specifics and more (e.g., additional hyperparameter settings) can be found in Appendices \ref{appendix:hyperparams} and \ref{appendix:templates}. 

\paragraph{Baselines.}
To benchmark FIRE's performance, we consider eight relevant and recent baselines: standard feedback-based on-policy self-distillation (Full-Context), Sample Routed Policy Optimization (SRPO) \citep{li2026unifying}, Trust Region Policy Distillation (TOP-D) \citep{xie2026trust},  Trust Region On-Policy Distillation (TrOPD) \citep{xing2026trust}, DemoPSD \citep{li2026demopsd}, Signal-Calibrated On-Policy Distillation Enhancement (SCOPE) \citep{zheng2026scope}, Veto \citep{jang2026stable}, and On-Policy Supervised Fine-tuning (On-Policy SFT) \citep{zhao2026policy}. We consider baselines that adopt approaches adjacent to FIRE to better quantify FIRE's performance differences. For example, TOP-D and TrOPD operate in a trust region similar to FIRE's Fisher-informed radius, while SCOPE conducts correct-incorrect branch self-distillation and SFT routing. DemoPSD similarly relies on calibration between the student and teacher distributions. Moreover, the baselines cover both self-distillation and regular on-policy distillation-based methods to better emphasize FIRE's advantages in a specifically self-distillation-based setting. More detailed specifics on how FIRE's methodology differs from the baselines can be found in Appendix \ref{appendix:novelty_info}.

\paragraph{Hardware and Additional Results.}
All experiments were conducted on a server with an NVIDIA A100-40GB GPU, utilizing the PyTorch \citep{paszke2019pytorch} and HuggingFace \citep{jain2022hugging} libraries. Additional experimental results, such as analysis on a larger model, ablation studies, etc. can be found in Appendix \ref{appendix:results}.

\begin{table*}[t]
\centering
\caption{Final accuracy (\%) across Qwen2.5-1.5B-Instruct and Qwen2.5-3B-Instruct and their associated datasets over three runs. FIRE exhibits superior performance to baselines across multiple datasets and model sizes.}
\label{tab:main_results}
\resizebox{\textwidth}{!}{%
\begin{tabular}{llcccc}
\toprule
\textbf{Model} & \textbf{Algorithm} & \textbf{GSM8K} & \textbf{ASDiv} &
\textbf{AQuA-RAT} & \textbf{Avg.} \\
\midrule

\multirow{9}{*}{\textbf{Qwen2.5-1.5B-Instruct}}
& Full Context   & $52.43{\pm}2.10$ & $76.04{\pm}3.25$ & $53.47{\pm}1.67$ & 60.65 \\
& On-Policy SFT  & $50.35{\pm}3.35$ & $77.26{\pm}0.30$ & $50.17{\pm}1.83$ & 59.26 \\
& Veto           & $50.35{\pm}5.22$ & $69.97{\pm}1.31$ & $54.86{\pm}2.57$ & 58.39 \\
& TOP-D          & $0.00{\pm}0.00$  & $0.00{\pm}0.00$  & $12.50{\pm}3.25$ & 4.17 \\
& TrOPD          & $45.66{\pm}4.84$ & $66.84{\pm}2.87$ & $48.78{\pm}2.10$ & 53.76 \\
& SRPO           & $59.55{\pm}1.97$ & $80.56{\pm}3.14$ & $49.65{\pm}3.39$ & 63.25 \\
& DemoPSD        & $55.21{\pm}5.49$ & $77.78{\pm}0.80$ & $54.51{\pm}3.55$ & 62.50 \\
& SCOPE          & $60.07{\pm}5.76$ & $77.60{\pm}2.27$ & $50.00{\pm}3.12$ & 62.56 \\
\rowcolor{blue!8}
& \textbf{FIRE (Ours)}
                 & $61.11{\pm}2.35$ & $81.94{\pm}1.31$ & $57.81{\pm}1.04$
                 & \textbf{66.96} \\

\midrule
\textbf{Model} & \textbf{Algorithm} & \textbf{ARC-Challenge} &
\textbf{WinoGrande} & \textbf{WiC} & \textbf{Avg.} \\
\midrule

\multirow{9}{*}{\textbf{Qwen2.5-3B-Instruct}}
& Full Context   & $82.12{\pm}2.17$ & $67.71{\pm}2.90$ & $63.54{\pm}0.52$ & 71.12 \\
& On-Policy SFT  & $83.51{\pm}3.05$ & $64.58{\pm}0.52$ & $63.54{\pm}0.52$ & 70.54 \\
& Veto           & $83.51{\pm}2.46$ & $65.28{\pm}2.87$ & $62.85{\pm}1.08$ & 70.54 \\
& TOP-D          & $83.33{\pm}1.88$ & $67.53{\pm}2.96$ & $55.21{\pm}4.51$ & 68.69 \\
& TrOPD          & $82.29{\pm}1.80$ & $64.76{\pm}2.41$ & $60.76{\pm}0.60$ & 69.27 \\
& SRPO           & $82.81{\pm}1.88$ & $63.72{\pm}2.46$ & $64.06{\pm}1.88$ & 70.20 \\
& DemoPSD        & $82.47{\pm}1.59$ & $67.36{\pm}4.05$ & $62.33{\pm}0.80$ & 70.72 \\
& SCOPE          & $84.38{\pm}2.71$ & $64.41{\pm}1.67$ & $65.97{\pm}0.80$ & 71.59 \\
\rowcolor{blue!8}
& \textbf{FIRE (Ours)}
                 & $86.11{\pm}2.46$ & $68.23{\pm}6.83$ & $64.76{\pm}2.10$
                 & \textbf{73.03} \\

\bottomrule
\end{tabular}%
}
\end{table*}

\subsection{Results}

\paragraph{Performance in Comparison to Baselines.} From the results presented in Table \ref{tab:main_results}, we note the superiority in performance of the proposed FIRE method over the baselines. FIRE achieves the highest average performance across the datasets on both Qwen2.5-1.5B-Instruct and Qwen2.5-3B-Instruct. This high performance exhibited by FIRE across all settings is in contrast to some of the baselines (e.g., TOP-D on ASDiv). This indicates that FIRE's methodology is more conducive to stable performance during training, in comparison to other on-policy self-distillation and regular on-policy distillation methods. We verify this trend holds for a significantly larger model in Appendix \ref{appendix_sec:14b}.

\begin{figure}[t]
\centering

\begin{subfigure}[t]{0.32\textwidth}
    \centering
    \includegraphics[width=\linewidth]{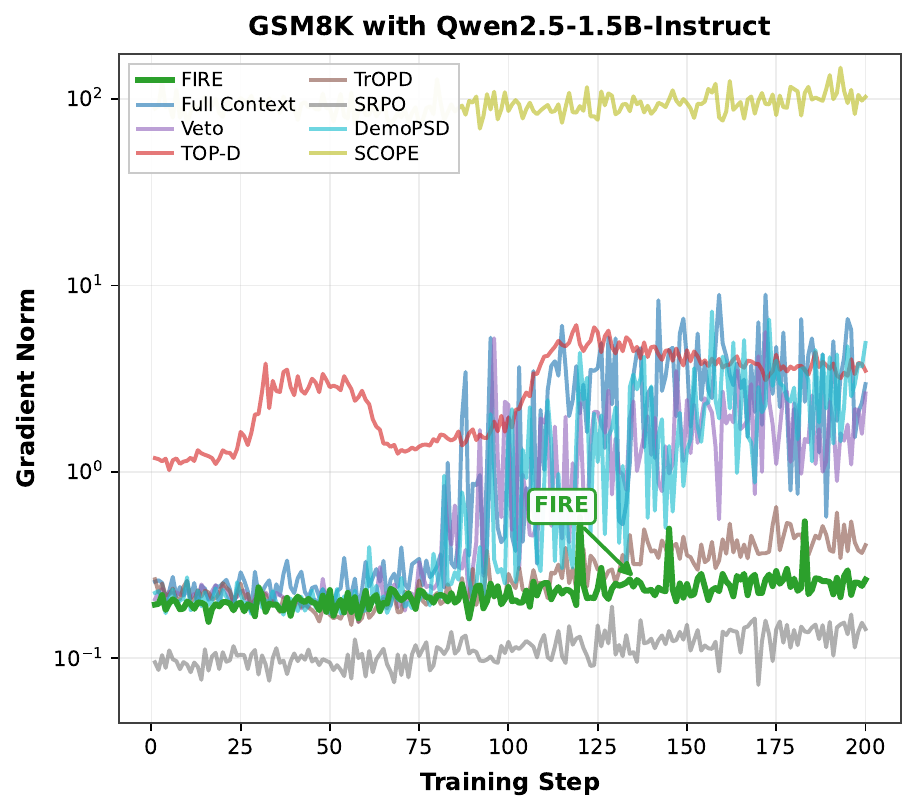}
    \caption*{GSM8K}
\end{subfigure}
\hfill
\begin{subfigure}[t]{0.32\textwidth}
    \centering
    \includegraphics[width=\linewidth]{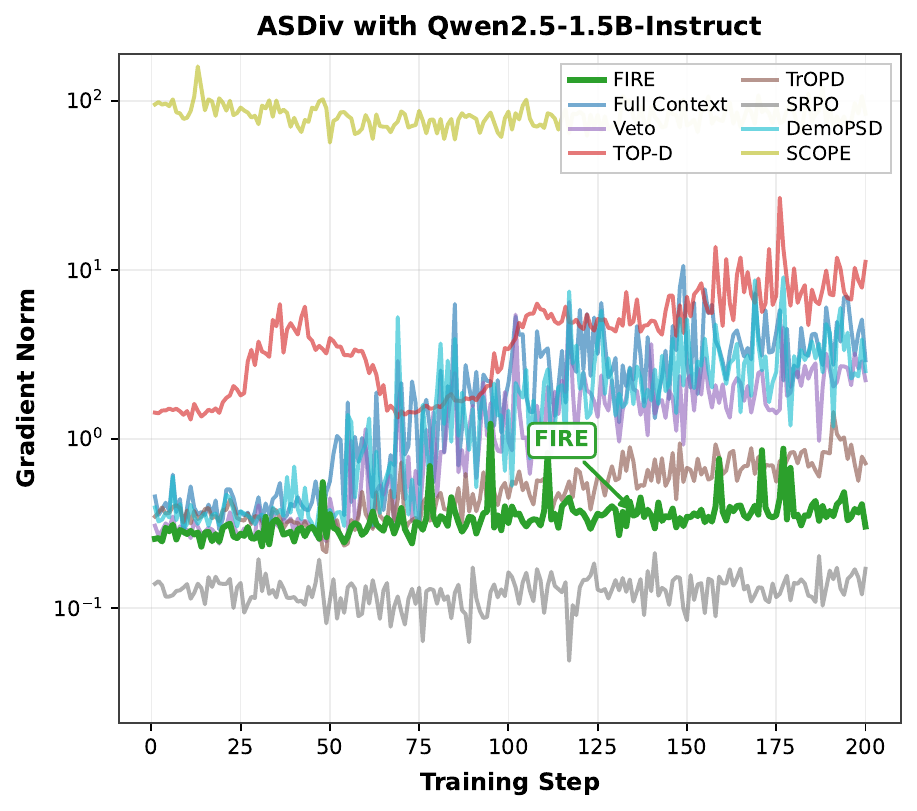}
    \caption*{ASDiv}
\end{subfigure}
\hfill
\begin{subfigure}[t]{0.32\textwidth}
    \centering
    \includegraphics[width=\linewidth]{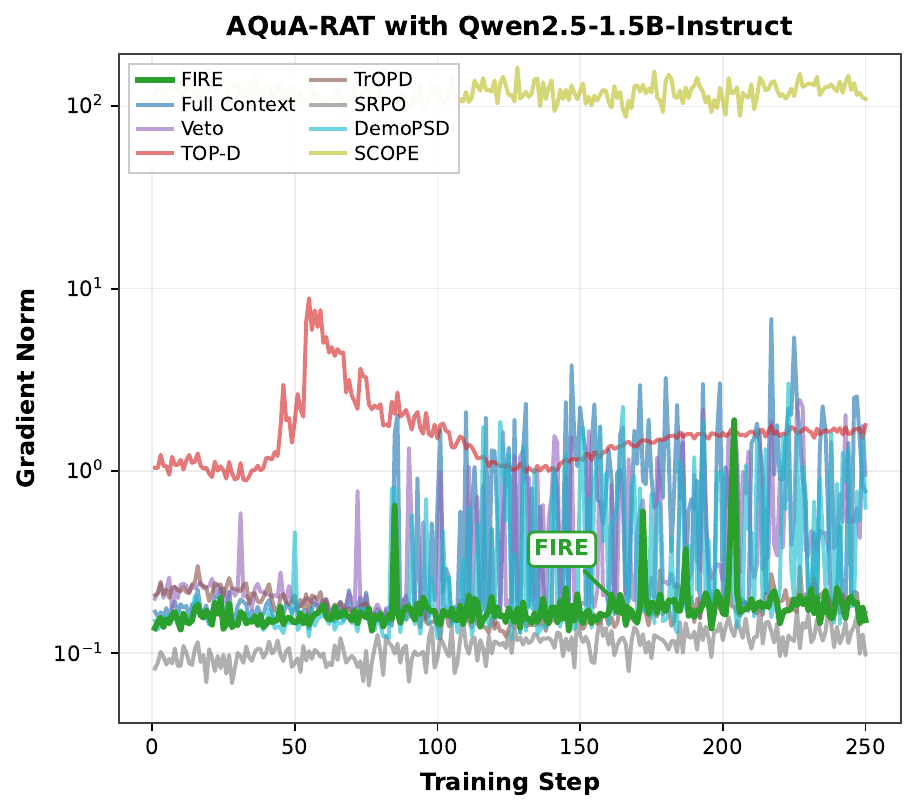}
    \caption*{AQuA-RAT}
\end{subfigure}

\vspace{0.5em}

{\small\textbf{(a) Qwen2.5-1.5B-Instruct}}

\vspace{0.5em}

\begin{subfigure}[t]{0.32\textwidth}
    \centering
    \includegraphics[width=\linewidth]{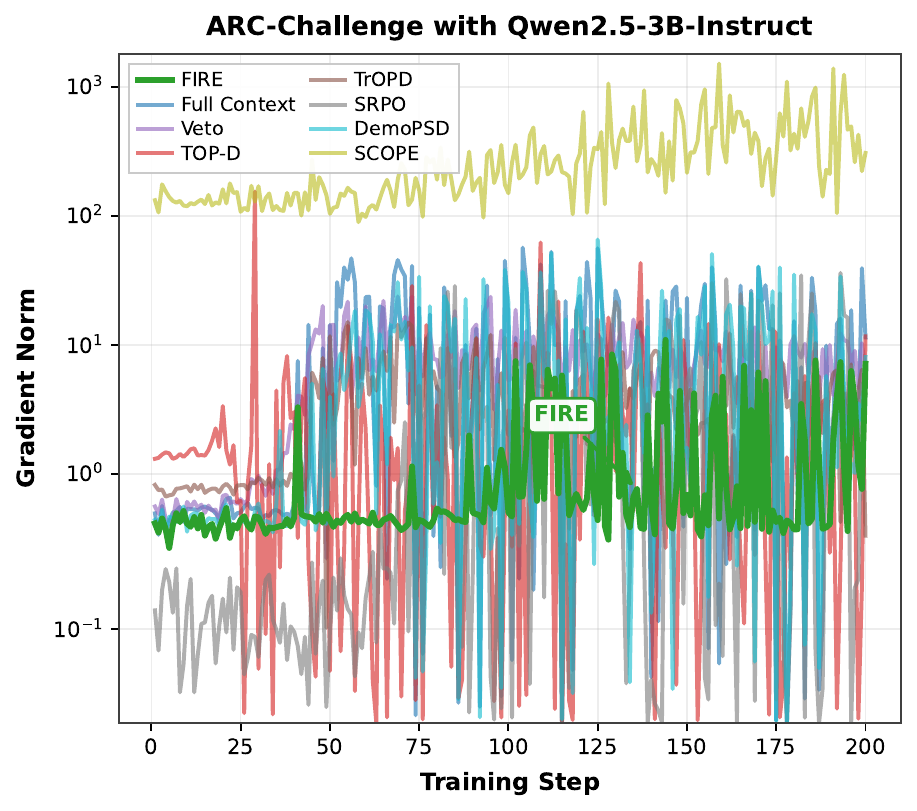}
    \caption*{ARC-Challenge}
\end{subfigure}
\hfill
\begin{subfigure}[t]{0.32\textwidth}
    \centering
    \includegraphics[width=\linewidth]{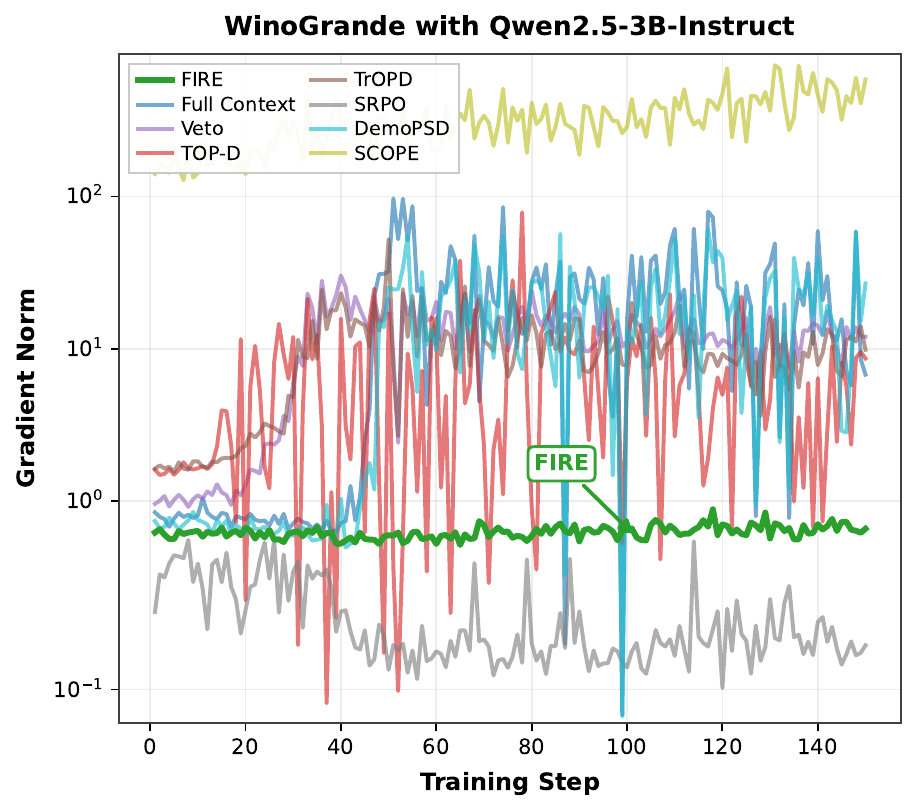}
    \caption*{WinoGrande}
\end{subfigure}
\hfill
\begin{subfigure}[t]{0.32\textwidth}
    \centering
    \includegraphics[width=\linewidth]{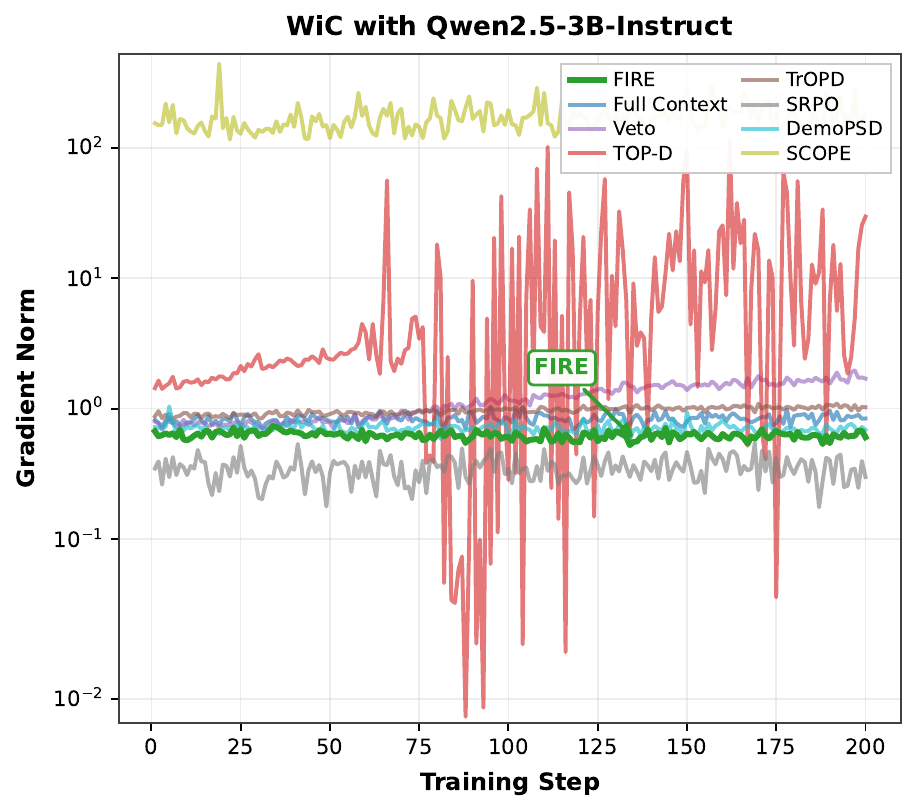}
    \caption*{WiC}
\end{subfigure}

\vspace{0.5em}

{\small\textbf{(b) Qwen2.5-3B-Instruct}}

\vspace{0.5em}

\caption{
Gradient norm on (a) Qwen2.5-1.5B-Instruct and (b) Qwen2.5-3B-Instruct. The plotted values show the gradient norms before clipping. During training, gradient clipping constrains the norm to $1.0$. The gradient norms are presented on a log-scale. FIRE exhibits much more stable training dynamics, with generally lower magnitude and less erratic oscillation.
}
\label{fig:main_grad_norms}

\end{figure}

\paragraph{Gradient Norm Stability.}
In Figure \ref{fig:main_grad_norms}, we examine whether FIRE's gradient-aware training induces more stable optimization dynamics. Although gradient clipping is applied to all methods during training, we report the \textit{pre-clipping} gradient norms, i.e., the norms before clipping. Clipping can prevent individual updates from becoming excessively large, but it can also mask underlying optimization instability. We exclude On-Policy SFT from this analysis because it does not perform distillation.

Across the evaluated settings, FIRE exhibits substantially more stable pre-clipping gradient norms than the distillation baselines. In particular, several methods (e.g., Veto and DemoPSD) develop highly variable gradients as training progresses, despite being subject to the same clipping procedure. In contrast, FIRE maintains nearly constant gradients on five of the six datasets. ARC-Challenge is the only setting in which FIRE exhibits oscillation. Importantly, however, this behavior emerges later in training and remains considerably smaller in magnitude (log-scale plotting) than the fluctuations observed for several baselines. SRPO is the only other method that demonstrates consistent gradient stability, but this stability comes at the cost of a substantially more expensive training procedure, as shown in Appendix \ref{appendix:tokens_gen_exp_section}.

These results provide evidence that FIRE improves the \textit{underlying optimization dynamics}, rather than relying on gradient clipping to suppress otherwise unstable updates. While clipping bounds the magnitude of the update, it does not make the training trajectories intrinsically stable. FIRE instead produces gradients that remain well-behaved even before this safeguard is applied, supporting the idea of its gradient-aware mechanism in stabilizing self-distillation.

\begin{figure}[t]
\centering

\begin{subfigure}[t]{0.3\textwidth}
    \centering
    \includegraphics[width=\linewidth]{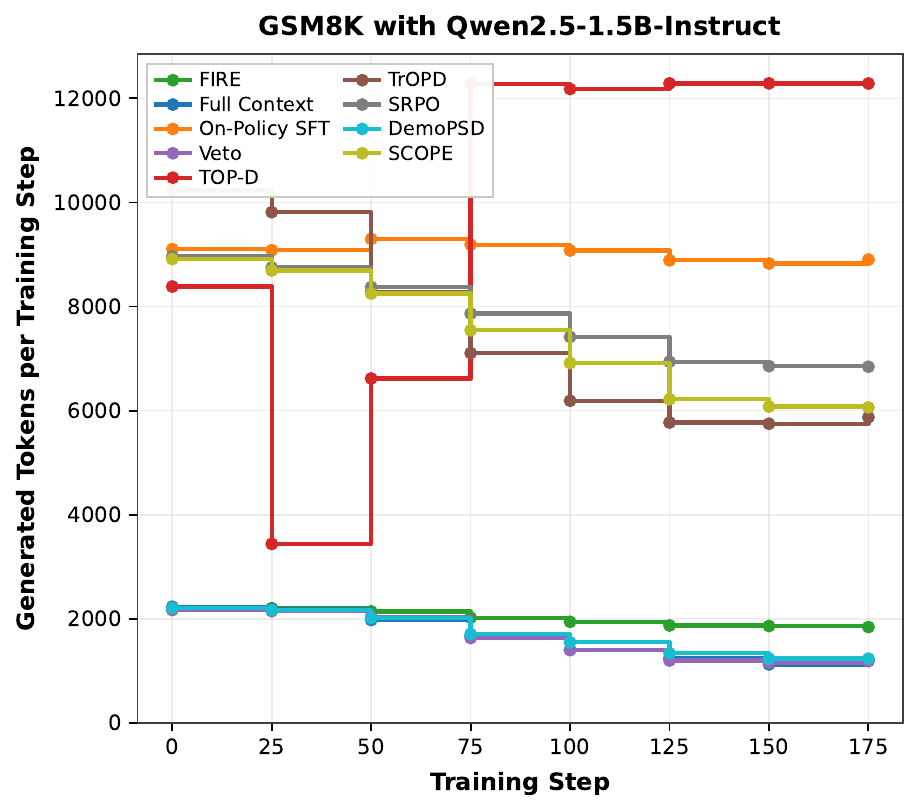}
    \caption*{GSM8K}
\end{subfigure}
\hfill
\begin{subfigure}[t]{0.3\textwidth}
    \centering
    \includegraphics[width=\linewidth]{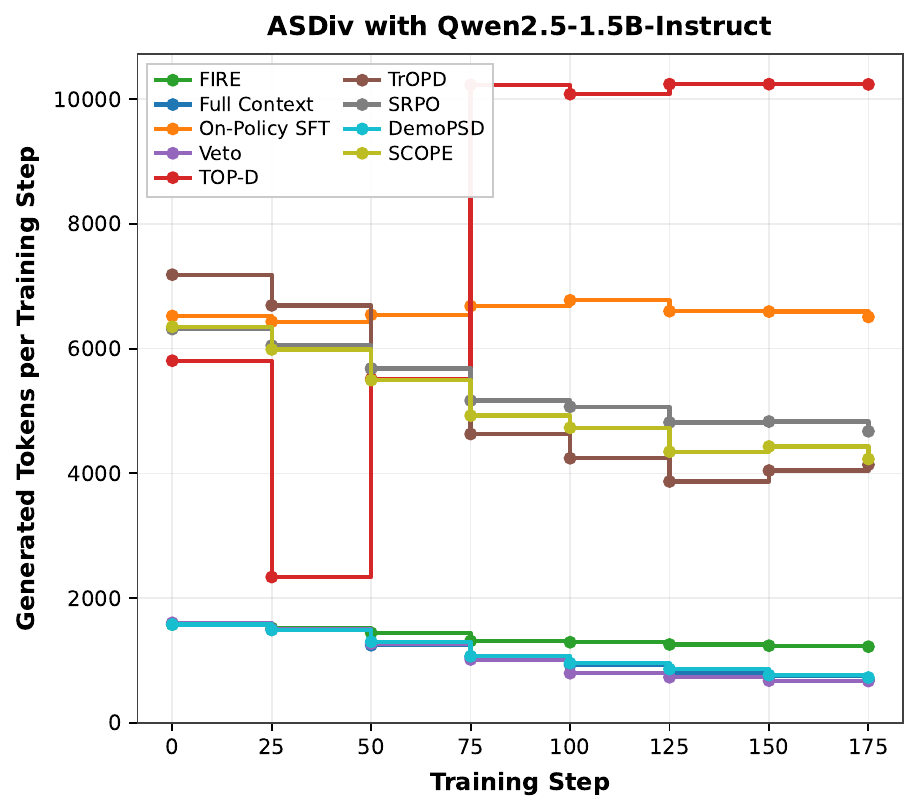}
    \caption*{ASDiv}
\end{subfigure}
\hfill
\begin{subfigure}[t]{0.3\textwidth}
    \centering
    \includegraphics[width=\linewidth]{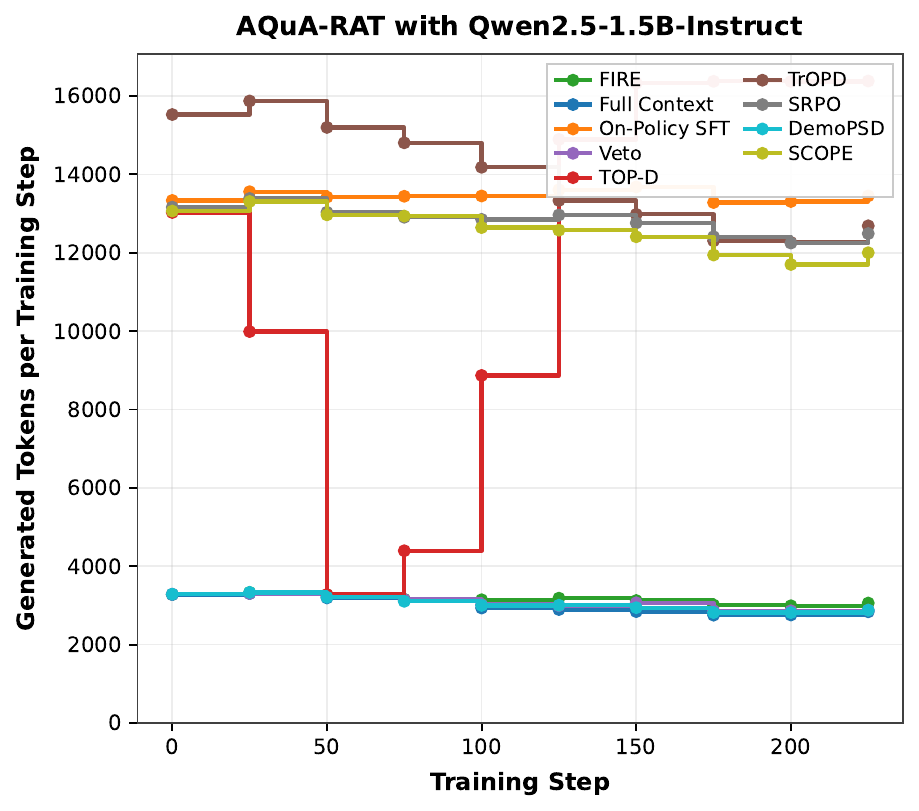}
    \caption*{AQuA-RAT}
\end{subfigure}

\vspace{0.5em}

{\small\textbf{(a) Qwen2.5-1.5B-Instruct}}

\vspace{0.5em}

\begin{subfigure}[t]{0.3\textwidth}
    \centering
    \includegraphics[width=\linewidth]{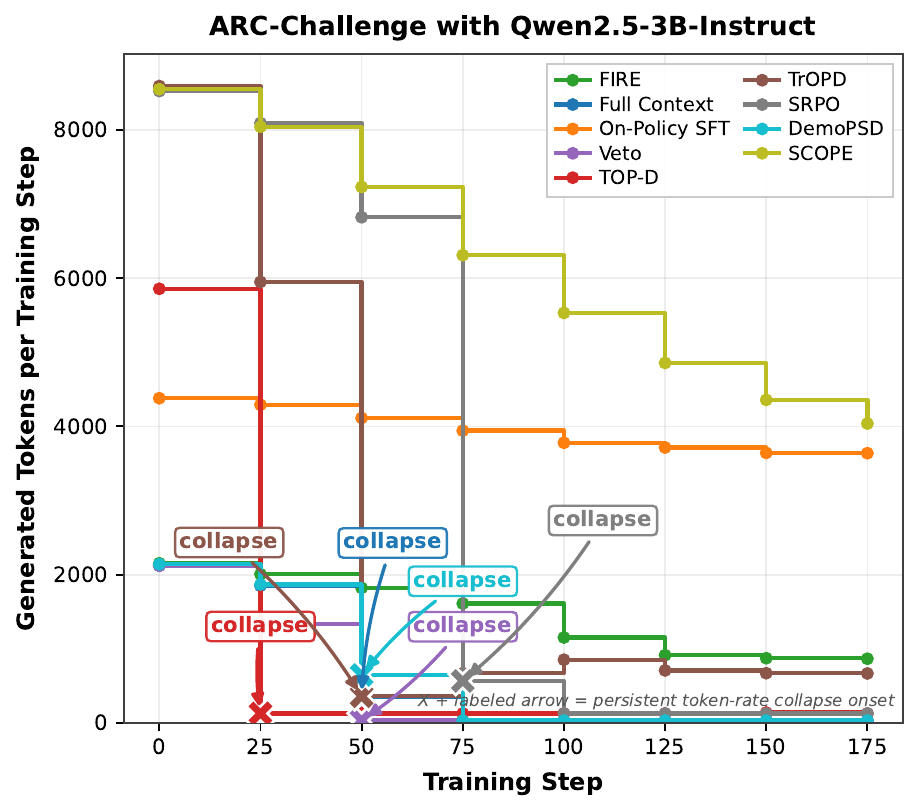}
    \caption*{ARC-Challenge}
\end{subfigure}
\hfill
\begin{subfigure}[t]{0.3\textwidth}
    \centering
    \includegraphics[width=\linewidth]{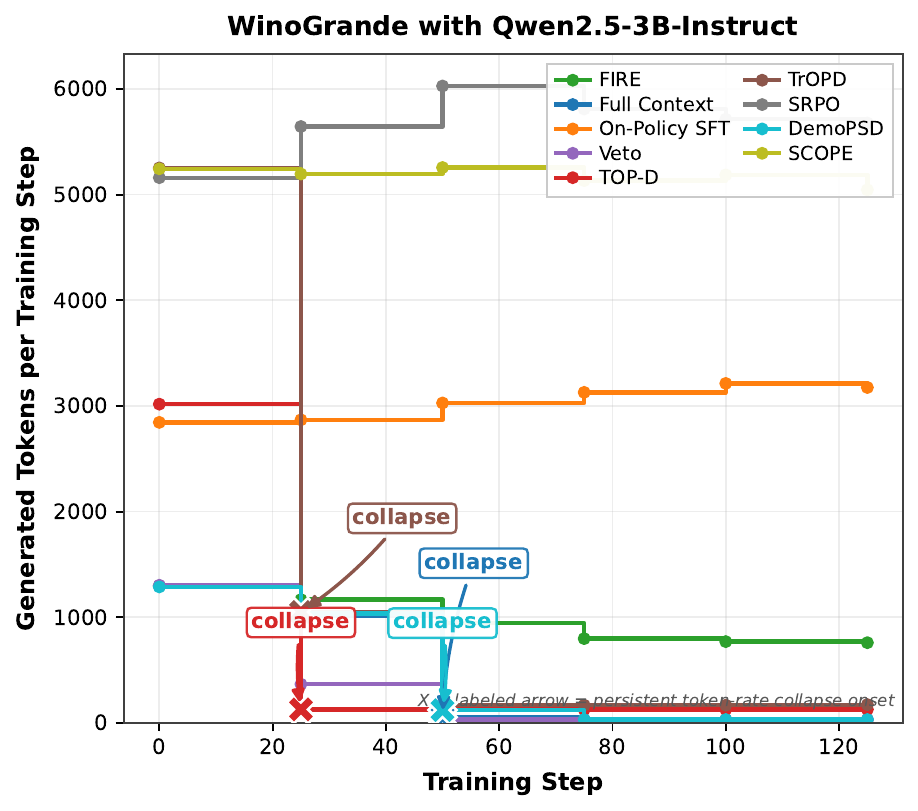}
    \caption*{WinoGrande}
\end{subfigure}
\hfill
\begin{subfigure}[t]{0.3\textwidth}
    \centering
    \includegraphics[width=\linewidth]{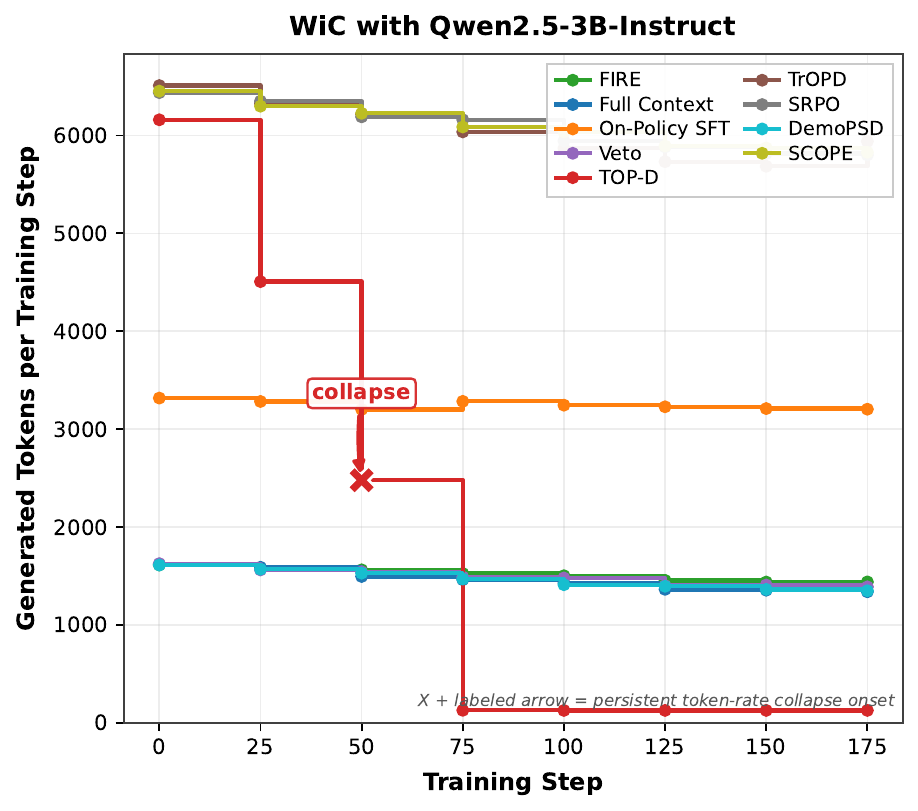}
    \caption*{WiC}
\end{subfigure}

\vspace{0.5em}

{\small\textbf{(b) Qwen2.5-3B-Instruct}}


\caption{
Number of auto-regressively generated tokens at each training step for each algorithm on (a) Qwen2.5-1.5B-Instruct and (b) Qwen2.5-3B-Instruct. FIRE's response length remains more resistant to collapse during training, maintaining its reasoning steps, in contrast to most baselines.}
\label{fig:main_gen_tokens}

\end{figure}

\paragraph{Number of Tokens Generated per Step.}
We next examine the number of tokens generated at each training step in Fig. \ref{fig:main_gen_tokens}, which provides another view of training stability. While Fig. \ref{fig:main_grad_norms} characterizes stability at the optimization level, generation length reveals whether instability manifests in the model's output behavior. In particular, a sudden and sustained reduction in generated tokens can indicate \textit{response collapse}, where training drives the model toward increasingly short responses.

As seen in Fig. \ref{fig:main_gen_tokens}, FIRE is relatively robust to this behavior. On Qwen2.5-1.5B-Instruct, its generation lengths remain relatively steady across all three datasets, while on ARC-Challenge and WinoGrande with Qwen2.5-3B-Instruct, FIRE exhibits only a gradual decline rather than an abrupt collapse. In contrast, several baselines (e.g., TOP-D, TrOPD, Full Context) exhibit sharp falls in response length during training. On-Policy SFT is the only baseline exhibiting similar characteristics, but requires more generated tokens, making it computationally inefficient in comparison to FIRE.

Overall, both Figs. \ref{fig:main_grad_norms} and \ref{fig:main_gen_tokens} show that FIRE's stability is not limited to the gradient norms but is simultaneously exhibited in response outputs. FIRE simultaneously maintains controlled pre-clipping gradients and avoids the abrupt response-length collapse exhibited by several baselines. Importantly, we additionally note the collapse correlates closely with the onset of increased gradient instability in Fig. \ref{fig:main_grad_norms}. 

\section{Conclusion}
In this paper, we introduced FIRE (\textbf{F}isher-\textbf{I}nformed \textbf{RE}calibration), a novel methodology for \textit{stabilizing feedback-based on-policy self-distillation in LLMs}. FIRE routes correct and incorrect on-policy responses through dual-branching, whereby correct responses are trained via Fisher-informed re-weighted on-policy SFT, while incorrect responses retain feedback-conditioned self-distillation with attribution-aware teacher recalibration. The attribution-aware recalibration identifies \textit{which direction} feedback should move the model, with the softmax Fisher trace primarily, in conjunction with gradient magnitudes and the learning rate, governing \emph{how far} updates should move the model across both branches. Across multiple datasets and model sizes, FIRE consistently achieves strong downstream performance after fine-tuning while exhibiting substantially more stable pre-clipping gradient norms and greater resistance to response-length collapse than existing baselines. Potential future work could explore more efficient methodologies for characterizing leave-one-field-out counterfactuals when the number of feedback fields is large.

\subsection*{AI use statement}

In this work, we used generative AI tools for improving clarity of the manuscript (paraphrasing, rewording, partial creation of Fig. \ref{fig:FIRE_method}, and creation of tables in the main text and Appendix). Moreover, it was used as an aid for the derivations in Appendix \ref{app:fire_details} and code implementation (e.g., implementation of proposed methodology and baselines). Research ideation of the proposed methodology was derived from the authors, and the rest of the required disclosure tasks (e.g., generation of new synthetic datasets) are not applicable to this work. We have reviewed all AI-assisted work. For example, we have confirmed the accuracy of the derivations of Appendix \ref{app:fire_details} and have checked the accuracy of the re-worded text. We take responsibility for the final content of this work, including text, claims or artifacts produced with the aid of generative AI.

\subsection*{Reproducibility statement}

We have provided the implementation of the FIRE methodology at \textcolor{blue}{\href{https://github.com/lee3296/FIRE}{this link}}, including configuration files. Moreover, details on hyperparameter and prompt template settings can be found in Sec. \ref{sec:experiments}, Appendix \ref{appendix:hyperparams}, and Appendix \ref{appendix:templates}. The pseudocode of the FIRE method can additionally be found in Appendix \ref{appendix:pseudocode}. Lastly, the derivations for the equations in Sec. \ref{sec:fire} can be found in full in Appendix \ref{app:fire_details}.



\bibliography{iclr2027_conference}

@article{li2026unifying,
  title={Unifying group-relative and self-distillation policy optimization via sample routing},
  author={Li, Gengsheng and Yang, Tianyu and Fang, Junfeng and Song, Mingyang and Zheng, Mao and Guo, Haiyun and Zhang, Dan and Wang, Jinqiao and Chua, Tat-Seng},
  journal={arXiv preprint arXiv:2604.02288},
  year={2026}
}

@article{zheng2026scope,
  title={Scope: Signal-calibrated on-policy distillation enhancement with dual-path adaptive weighting},
  author={Zheng, Binbin and Ma, Xing and Liang, Yiheng and Ruan, Jingqing and Fu, Xiaoliang and Lin, Kepeng and Zhu, Benchang and Zeng, Ke and Cai, Xunliang},
  journal={arXiv preprint arXiv:2604.10688},
  year={2026}
}

@article{li2026demopsd,
  title={DemoPSD: Disagreement-Modulated Policy Self-Distillation},
  author={Li, Yunhe and Shi, Hao and Liu, Wenhao and Ruan, Mengzhe and Hou, Hanxu and Dai, Zhongxiang and Qiu, Shuang and Song, Linqi},
  journal={arXiv preprint arXiv:2607.02502},
  year={2026}
}

@article{minaee2024large,
  title={Large language models: A survey},
  author={Minaee, Shervin and Mikolov, Tomas and Nikzad, Narjes and Chenaghlu, Meysam and Socher, Richard and Amatriain, Xavier and Gao, Jianfeng},
  journal={arXiv preprint arXiv:2402.06196},
  year={2024}
}

@article{bommasani2021opportunities,
  title={On the opportunities and risks of foundation models},
  author={Bommasani, Rishi and Hudson, Drew A and Adeli, Ehsan and Altman, Russ and Arora, Simran and von Arx, Sydney and Bernstein, Michael S and Bohg, Jeannette and Bosselut, Antoine and Brunskill, Emma and others},
  journal={arXiv preprint arXiv:2108.07258},
  year={2021}
}

@article{grattafiori2024llama,
  title={The llama 3 herd of models},
  author={Grattafiori, Aaron and Dubey, Abhimanyu and Jauhri, Abhinav and Pandey, Abhinav and Kadian, Abhishek and Al-Dahle, Ahmad and Letman, Aiesha and Mathur, Akhil and Schelten, Alan and Vaughan, Alex and others},
  journal={arXiv preprint arXiv:2407.21783},
  year={2024}
}

@article{yang2025qwen3,
  title={Qwen3 technical report},
  author={Yang, An and Li, Anfeng and Yang, Baosong and Zhang, Beichen and Hui, Binyuan and Zheng, Bo and Yu, Bowen and Gao, Chang and Huang, Chengen and Lv, Chenxu and others},
  journal={arXiv preprint arXiv:2505.09388},
  year={2025}
}

@article{hu2021lora,
  title={Lora: Low-rank adaptation of large language models},
  author={Hu, Edward J and Shen, Yelong and Wallis, Phillip and Allen-Zhu, Zeyuan and Li, Yuanzhi and Wang, Shean and Wang, Lu and Chen, Weizhu},
  journal={arXiv preprint arXiv:2106.09685},
  year={2021}
}

@article{zhang2025parameter,
  title={Parameter-efficient fine-tuning for foundation models},
  author={Zhang, Dan and Feng, Tao and Xue, Lilong and Wang, Yuandong and Dong, Yuxiao and Tang, Jie},
  journal={arXiv preprint arXiv:2501.13787},
  year={2025}
}

@article{lee2026self,
  title={Self-Play Enhancement via Advantage-Weighted Refinement in Online Federated LLM Fine-Tuning with Real-Time Feedback},
  author={Lee, Seohyun and Fang, Wenzhi and Han, Dong-Jun and Hosseinalipour, Seyyedali and Brinton, Christopher G},
  journal={arXiv preprint arXiv:2605.07977},
  year={2026}
}

@article{hubotter2026reinforcement,
  title={Reinforcement learning via self-distillation},
  author={H{\"u}botter, Jonas and L{\"u}beck, Frederike and Behric, Lejs and Baumann, Anton and Bagatella, Marco and Marta, Daniel and Hakimi, Ido and Shenfeld, Idan and Buening, Thomas Kleine and Guestrin, Carlos and others},
  journal={arXiv preprint arXiv:2601.20802},
  year={2026}
}

@article{kullback1951information,
  title={On information and sufficiency},
  author={Kullback, Solomon and Leibler, Richard A},
  journal={The annals of mathematical statistics},
  volume={22},
  number={1},
  pages={79--86},
  year={1951},
  publisher={JSTOR}
}

@article{zhao2026self,
  title={Self-distilled reasoner: On-policy self-distillation for large language models},
  author={Zhao, Siyan and Xie, Zhihui and Liu, Mengchen and Huang, Jing and Pang, Guan and Chen, Feiyu and Grover, Aditya},
  journal={arXiv preprint arXiv:2601.18734},
  year={2026}
}

@article{fang2025federated,
  title={Federated sketching lora: On-device collaborative fine-tuning of large language models},
  author={Fang, Wenzhi and Han, Dong-Jun and Yuan, Liangqi and Hosseinalipour, Seyyedali and Brinton, Christopher G},
  journal={arXiv preprint arXiv:2501.19389},
  year={2025}
}

@article{cobbe2021gsm8k,
  title={Training Verifiers to Solve Math Word Problems},
  author={Cobbe, Karl and Kosaraju, Vineet and Bavarian, Mohammad and Chen, Mark and Jun, Heewoo and Kaiser, Lukasz and Plappert, Matthias and Tworek, Jerry and Hilton, Jacob and Nakano, Reiichiro and Hesse, Christopher and Schulman, John},
  journal={arXiv preprint arXiv:2110.14168},
  year={2021}
}

@article{ling2017program,
  title={Program induction by rationale generation: Learning to solve and explain algebraic word problems},
  author={Ling, Wang and Yogatama, Dani and Dyer, Chris and Blunsom, Phil},
  journal={ACL},
  year={2017}
}

@inproceedings{miao-etal-2020-diverse,
  title={A Diverse Corpus for Evaluating and Developing English Math Word Problem Solvers},
  author={Miao, Shen-yun and Liang, Chao-Chun and Su, Keh-Yih},
  booktitle={Proceedings of the 58th Annual Meeting of the Association for Computational Linguistics},
  pages={975--984},
  year={2020}
}

@article{allenai:arc,
      author    = {Peter Clark  and Isaac Cowhey and Oren Etzioni and Tushar Khot and
                    Ashish Sabharwal and Carissa Schoenick and Oyvind Tafjord},
      title     = {Think you have Solved Question Answering? Try ARC, the AI2 Reasoning Challenge},
      journal   = {arXiv:1803.05457v1},
      year      = {2018},
}

@article{wang2019superglue,
  title={Super{GLUE}: A Stickier Benchmark for General-Purpose Language Understanding Systems},
  author={Alex Wang and Yada Pruksachatkun and Nikita Nangia and Amanpreet Singh and Julian Michael and Felix Hill and Omer Levy and Samuel R. Bowman},
  journal={arXiv preprint 1905.00537},
  year={2019}
}

@inproceedings{pilehvar2018wic,
  title={{WiC}: The Word-in-Context Dataset for Evaluating Context-Sensitive Meaning Representations},
  author={Pilehvar, Mohammad Taher and Camacho-Collados, Jose},
  booktitle={Proceedings of NAACL-HLT},
  year={2019}
}

@article{loshchilov2017decoupled,
  title={Decoupled weight decay regularization},
  author={Loshchilov, Ilya and Hutter, Frank},
  journal={arXiv preprint arXiv:1711.05101},
  year={2017}
}

@misc{qwen2.5,
    title = {Qwen2.5: A Party of Foundation Models},
    url = {https://qwenlm.github.io/blog/qwen2.5/},
    author = {{Qwen Team}},
    month = {September},
    year = {2024}
}

@article{xie2026trust,
  title={Trust Region Policy Distillation},
  author={Xie, Zhengpeng and Zhang, Li Lyna and Xie, Zeke and Yang, Mao},
  journal={arXiv preprint arXiv:2607.04751},
  year={2026}
}

@article{xing2026trust,
  title={Trust Region On-Policy Distillation},
  author={Xing, Xingrun and Wang, Haoqing and Gao, Boyan and Li, Ziheng and Tang, Yehui},
  journal={arXiv preprint arXiv:2606.01249},
  year={2026}
}

@inproceedings{jang2026stable,
  title={Stable on-policy distillation through adaptive target reformulation},
  author={Jang, Ijun and Yeom, Jewon and Yeo, Juan and Lim, Hyunggyu and Kim, Taesup},
  booktitle={Findings of the Association for Computational Linguistics: ACL 2026},
  pages={42217--42227},
  year={2026}
}

@article{zhao2026policy,
  title={On-policy supervised fine-tuning for efficient reasoning},
  author={Zhao, Anhao and Chen, Ziyang and Tong, Junlong and Fan, Yingqi and Ye, Fanghua and Li, Shuhao and Ma, Yunpu and Li, Wenjie and Shen, Xiaoyu},
  journal={arXiv preprint arXiv:2602.13407},
  year={2026}
}

@article{paszke2019pytorch,
  title={Pytorch: An imperative style, high-performance deep learning library},
  author={Paszke, Adam and Gross, Sam and Massa, Francisco and Lerer, Adam and Bradbury, James and Chanan, Gregory and Killeen, Trevor and Lin, Zeming and Gimelshein, Natalia and Antiga, Luca and others},
  journal={Advances in neural information processing systems},
  volume={32},
  year={2019}
}

@incollection{jain2022hugging,
  title={Hugging face},
  author={Jain, Shashank Mohan},
  booktitle={Introduction to transformers for NLP: With the hugging face library and models to solve problems},
  pages={51--67},
  year={2022},
  publisher={Springer}
}

@inproceedings{liu2024moe,
  title={When moe meets llms: Parameter efficient fine-tuning for multi-task medical applications},
  author={Liu, Qidong and Wu, Xian and Zhao, Xiangyu and Zhu, Yuanshao and Xu, Derong and Tian, Feng and Zheng, Yefeng},
  booktitle={Proceedings of the 47th international ACM SIGIR conference on research and development in information retrieval},
  pages={1104--1114},
  year={2024}
}

@inproceedings{li2021prefix,
  title={Prefix-tuning: Optimizing continuous prompts for generation},
  author={Li, Xiang Lisa and Liang, Percy},
  booktitle={Proceedings of the 59th annual meeting of the association for computational linguistics and the 11th international joint conference on natural language processing (volume 1: Long papers)},
  pages={4582--4597},
  year={2021}
}

@article{lee2025tap,
  title={TAP: Two-Stage Adaptive Personalization of Multi-task and Multi-Modal Foundation Models in Federated Learning},
  author={Lee, Seohyun and Fang, Wenzhi and Han, Dong-Jun and Hosseinalipour, Seyyedali and Brinton, Christopher G},
  journal={arXiv preprint arXiv:2509.26524},
  year={2025}
}

@article{hinton2015distilling,
  title={Distilling the knowledge in a neural network},
  author={Hinton, Geoffrey and Vinyals, Oriol and Dean, Jeff},
  journal={arXiv preprint arXiv:1503.02531},
  year={2015}
}

@inproceedings{phuong2019towards,
  title={Towards understanding knowledge distillation},
  author={Phuong, Mary and Lampert, Christoph},
  booktitle={International conference on machine learning},
  pages={5142--5151},
  year={2019},
  organization={PMLR}
}

@article{xu2024survey,
  title={A survey on knowledge distillation of large language models},
  author={Xu, Xiaohan and Li, Ming and Tao, Chongyang and Shen, Tao and Cheng, Reynold and Li, Jinyang and Xu, Can and Tao, Dacheng and Zhou, Tianyi},
  journal={arXiv preprint arXiv:2402.13116},
  year={2024}
}

@article{shenfeld2026self,
  title={Self-distillation enables continual learning},
  author={Shenfeld, Idan and Damani, Mehul and H{\"u}botter, Jonas and Agrawal, Pulkit},
  journal={arXiv preprint arXiv:2601.19897},
  year={2026}
}

@article{wang2024mmlu,
  title={Mmlu-pro: A more robust and challenging multi-task language understanding benchmark},
  author={Wang, Yubo and Ma, Xueguang and Zhang, Ge and Ni, Yuansheng and Chandra, Abhranil and Guo, Shiguang and Ren, Weiming and Arulraj, Aaran and He, Xuan and Jiang, Ziyan and others},
  journal={Advances in Neural Information Processing Systems},
  volume={37},
  pages={95266--95290},
  year={2024}
}

@article{shao2024deepseekmath,
  title={Deepseekmath: Pushing the limits of mathematical reasoning in open language models},
  author={Shao, Zhihong and Wang, Peiyi and Zhu, Qihao and Xu, Runxin and Song, Junxiao and Bi, Xiao and Zhang, Haowei and Zhang, Mingchuan and Li, YK and Wu, Yang and others},
  journal={arXiv preprint arXiv:2402.03300},
  year={2024}
}

@article{ouyang2022training,
  title={Training language models to follow instructions with human feedback},
  author={Ouyang, Long and Wu, Jeffrey and Jiang, Xu and Almeida, Diogo and Wainwright, Carroll and Mishkin, Pamela and Zhang, Chong and Agarwal, Sandhini and Slama, Katarina and Ray, Alex and others},
  journal={Advances in neural information processing systems},
  volume={35},
  pages={27730--27744},
  year={2022}
}

@article{sakaguchi2021winogrande,
  title={Winogrande: An adversarial winograd schema challenge at scale},
  author={Sakaguchi, Keisuke and Bras, Ronan Le and Bhagavatula, Chandra and Choi, Yejin},
  journal={Communications of the ACM},
  volume={64},
  number={9},
  pages={99--106},
  year={2021},
  publisher={ACM New York, NY, USA}
}
\bibliographystyle{iclr2027_conference}

\startcontents[sections]

\onecolumn
\newpage

\appendix

\begin{center}
    {\bfseries\Large Appendix}
\end{center}

\printcontents[sections]{l}{1}{\setcounter{tocdepth}{3}}

\newpage

\appendix


\section{Additional Experimental Results}\label{appendix:results}
Here, we present additional experimental results in addition to those presented in Sec. \ref{sec:experiments}. If hyperparameter settings differ for certain experiments, they will be stated in their corresponding subsections.

\subsection{Auto-regressively Generated Tokens during Training vs. Accuracy}\label{appendix:tokens_gen_exp_section}

\begin{figure}[htbp]
\centering

\begin{subfigure}[t]{0.32\textwidth}
    \centering
    \includegraphics[width=\linewidth]{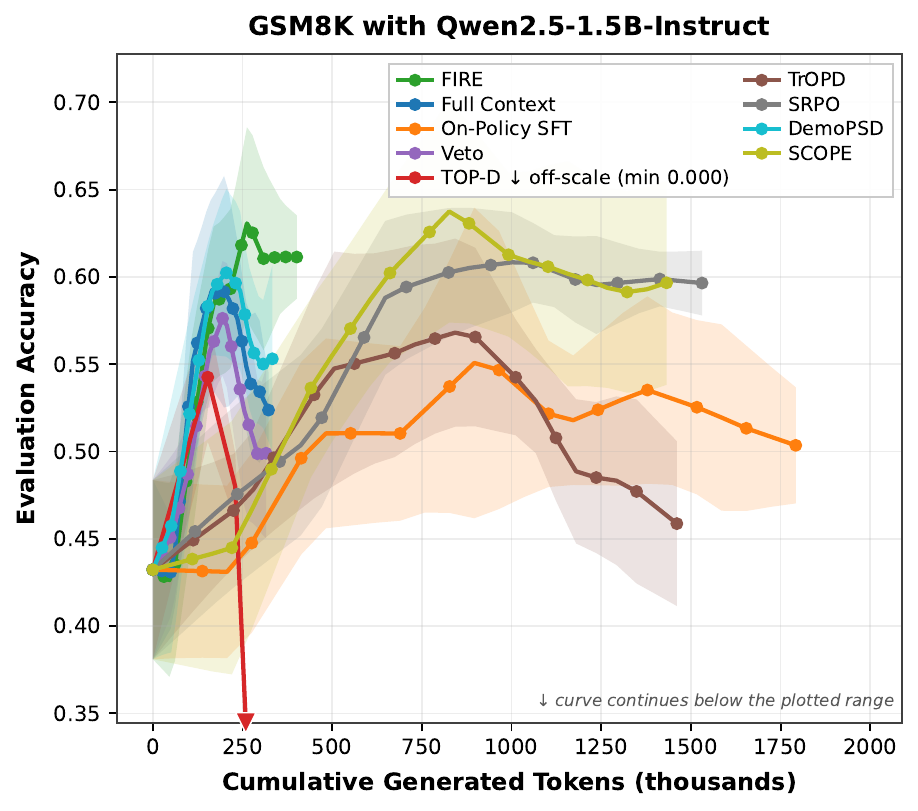}
    \caption*{GSM8K}
\end{subfigure}
\hfill
\begin{subfigure}[t]{0.32\textwidth}
    \centering
    \includegraphics[width=\linewidth]{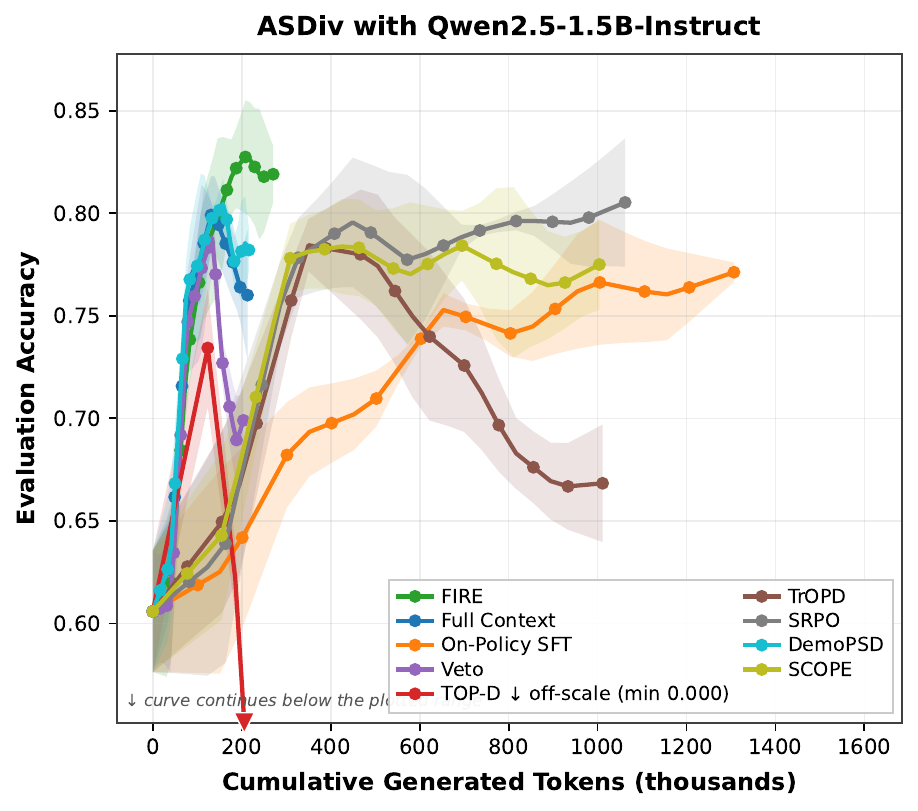}
    \caption*{ASDiv}
\end{subfigure}
\hfill
\begin{subfigure}[t]{0.32\textwidth}
    \centering
    \includegraphics[width=\linewidth]{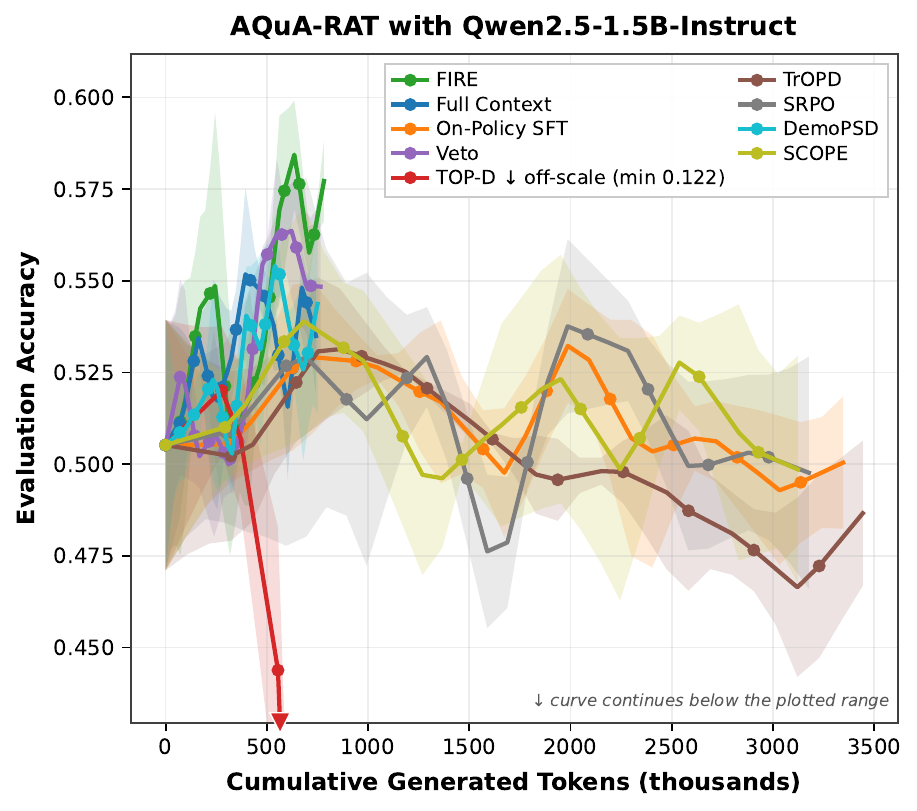}
    \caption*{AQuA-RAT}
\end{subfigure}

\vspace{0.5em}

{\small\textbf{(a) Qwen2.5-1.5B-Instruct}}

\vspace{0.75em}

\begin{subfigure}[t]{0.32\textwidth}
    \centering
    \includegraphics[width=\linewidth]{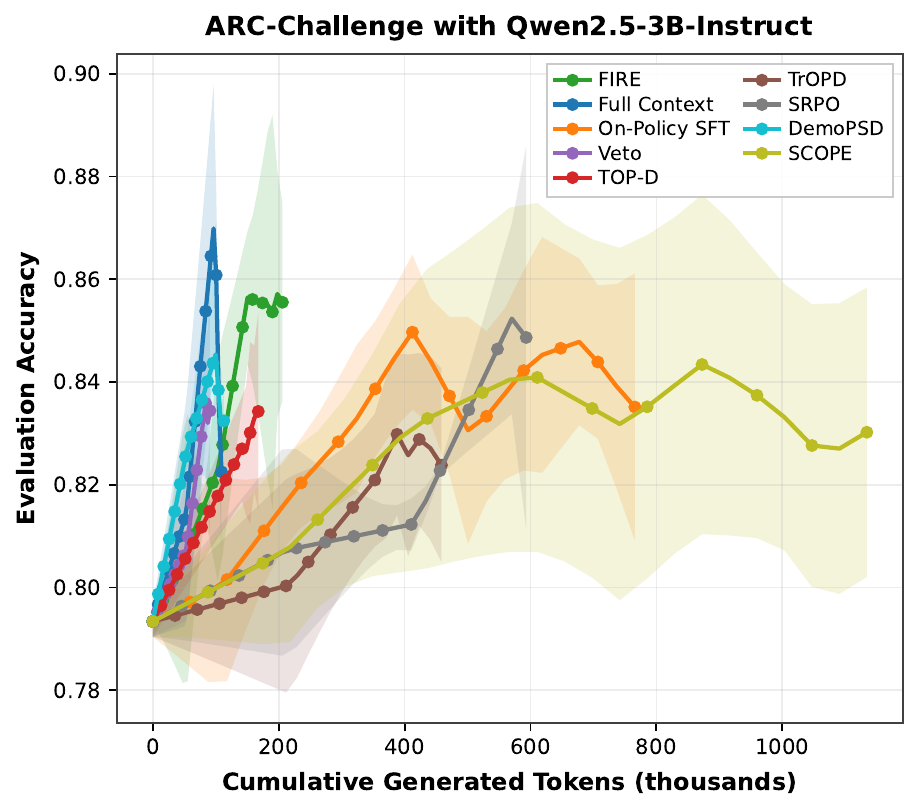}
    \caption*{ARC-Challenge}
\end{subfigure}
\hfill
\begin{subfigure}[t]{0.32\textwidth}
    \centering
    \includegraphics[width=\linewidth]{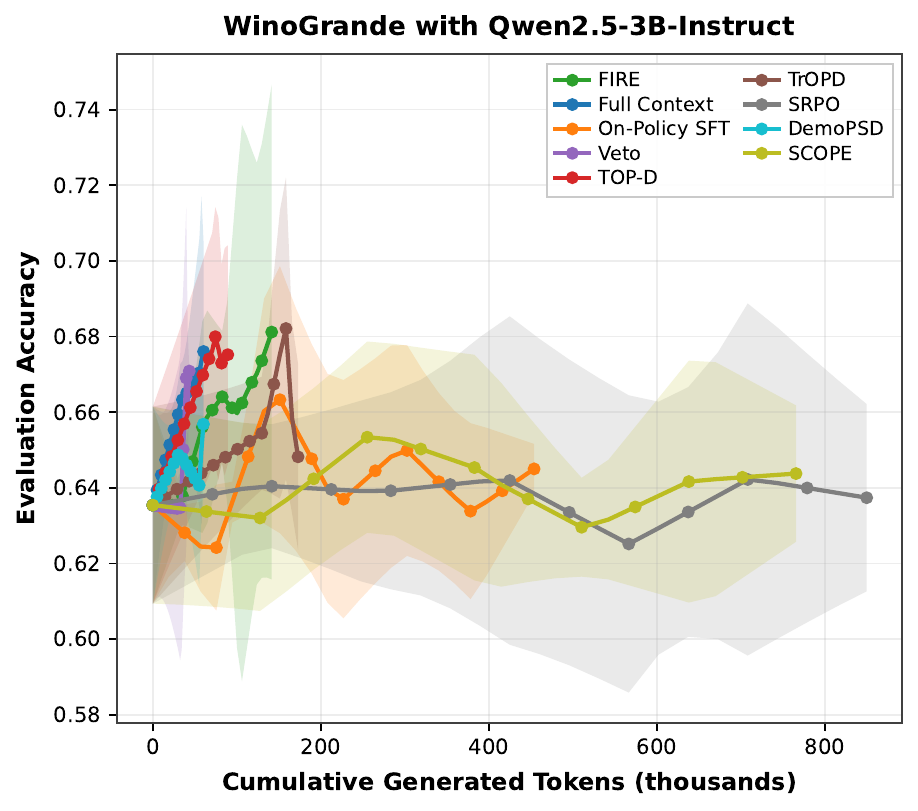}
    \caption*{WinoGrande}
\end{subfigure}
\hfill
\begin{subfigure}[t]{0.32\textwidth}
    \centering
    \includegraphics[width=\linewidth]{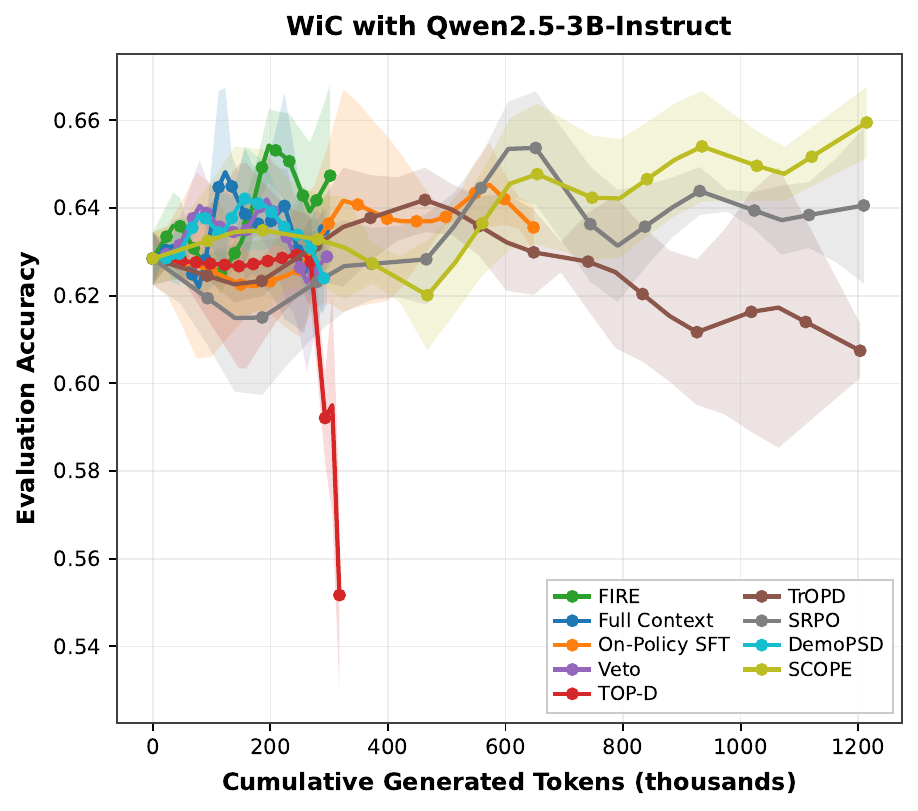}
    \caption*{WiC}
\end{subfigure}

\vspace{0.5em}

{\small\textbf{(b) Qwen2.5-3B-Instruct}}

\vspace{0.5em}

\caption{
Number of cumulatively auto-regressively generated tokens versus accuracy of each algorithm on (a) Qwen2.5-1.5B-Instruct and (b) Qwen2.5-3B-Instruct. FIRE offers the best tradeoff between training efficiency and accuracy.
}
\label{appendix_exp:gen_tokens_vs_acc}

\end{figure}

Here, we consider the number of cumulatively generated tokens auto-regressively versus test accuracy across FIRE and the baselines in Fig. \ref{appendix_exp:gen_tokens_vs_acc}. For the curves, each run's trajectory is linearly interpolated onto a shared grid over the range jointly covered by all runs for that method. From Fig. \ref{appendix_exp:gen_tokens_vs_acc}, we note that across five of the six datasets, FIRE offers the best tradeoff between training efficiency and accuracy. While SCOPE attains a higher final accuracy than FIRE on WiC, it is computationally more expensive to reach that point. Moreover, observations noted in Sec. \ref{sec:experiments}, specifically that FIRE induces better stability properties can also be seen in Fig. \ref{appendix_exp:gen_tokens_vs_acc} in comparison to SCOPE (e.g., continuous decrease in accuracy of SCOPE on AQuA-RAT). This is true for other datasets as well. For example, while TOP-D comes close to beating FIRE on WinoGrande, it is the worst performing method on all Qwen2.5-1.5B-Instruct-based datasets and WiC; it also lags behind on ARC-Challenge. Additional examples with similar observations include TrOPD, DemoPSD, and Full Context. Methods that exhibit more stability, such as SRPO or On-Policy SFT, are computationally much more expensive than FIRE. This is because methods such as SRPO rely on group calculations, which require generating multiple completions for one input sample.

\subsection{Training Time vs. Accuracy}\label{appendix_subsec:timevsacc}

\begin{figure}[t]
\centering

\begin{subfigure}[t]{0.32\textwidth}
    \centering
    \includegraphics[width=\linewidth]{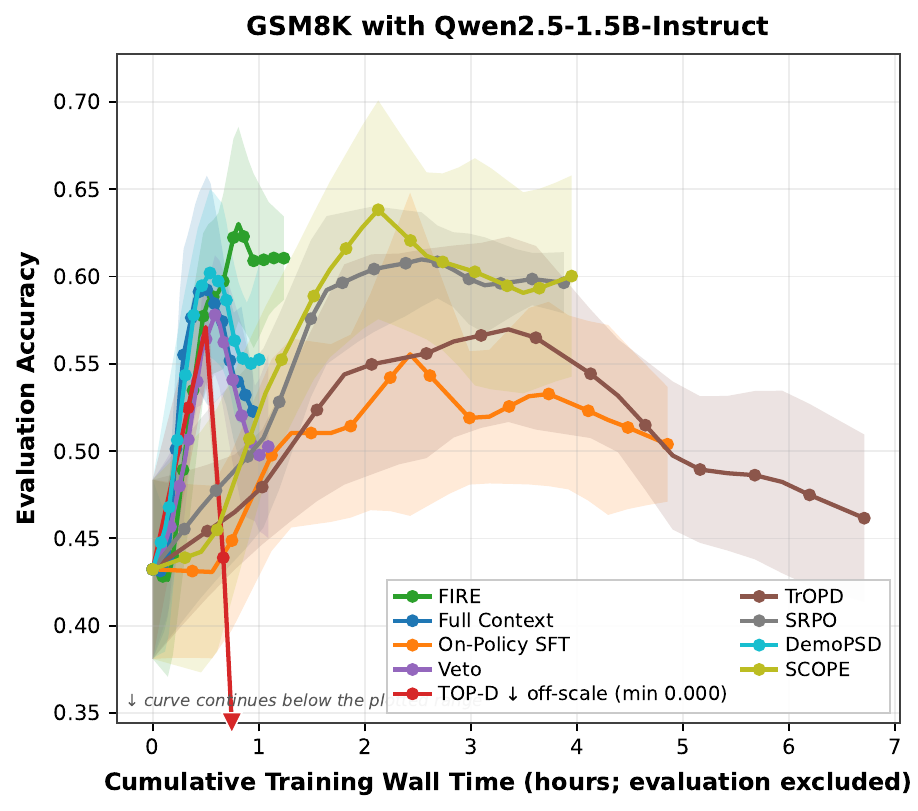}
    \caption*{GSM8K}
\end{subfigure}
\hfill
\begin{subfigure}[t]{0.32\textwidth}
    \centering
    \includegraphics[width=\linewidth]{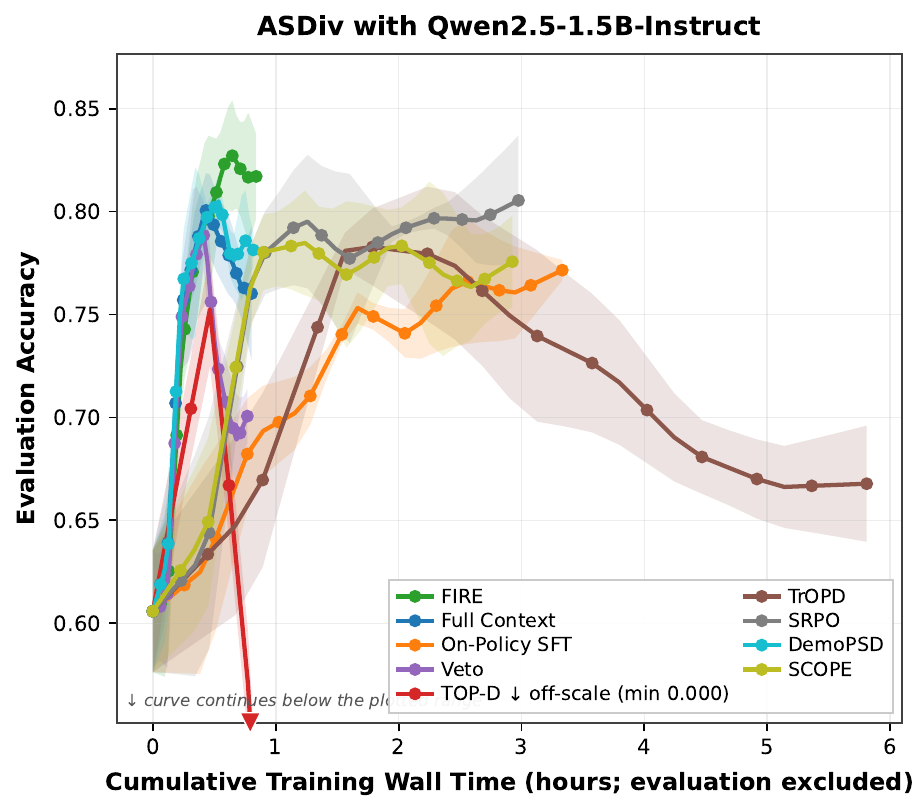}
    \caption*{ASDiv}
\end{subfigure}
\hfill
\begin{subfigure}[t]{0.32\textwidth}
    \centering
    \includegraphics[width=\linewidth]{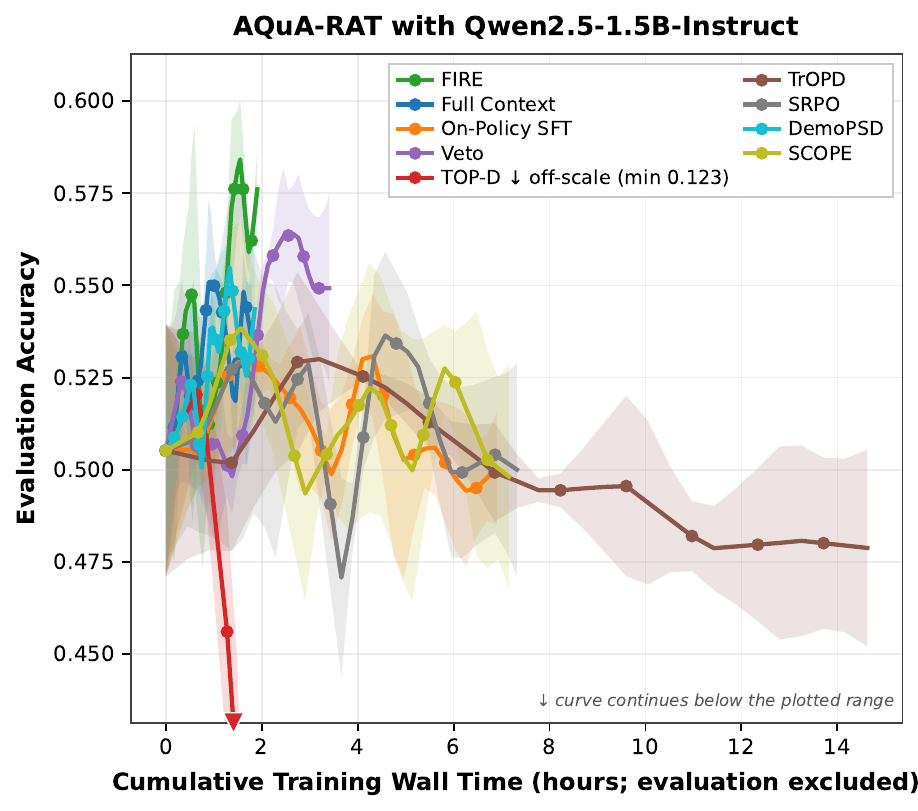}
    \caption*{AQuA-RAT}
\end{subfigure}

\vspace{0.5em}

{\small\textbf{(a) Qwen2.5-1.5B-Instruct}}

\vspace{0.75em}

\begin{subfigure}[t]{0.32\textwidth}
    \centering
    \includegraphics[width=\linewidth]{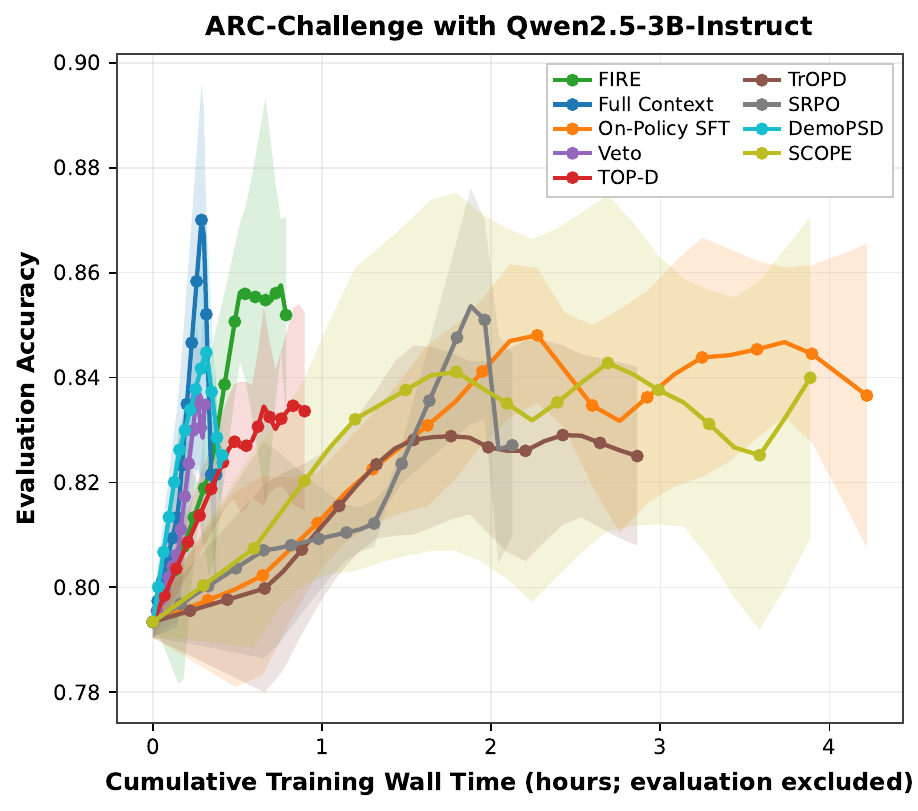}
    \caption*{ARC-Challenge}
\end{subfigure}
\hfill
\begin{subfigure}[t]{0.32\textwidth}
    \centering
    \includegraphics[width=\linewidth]{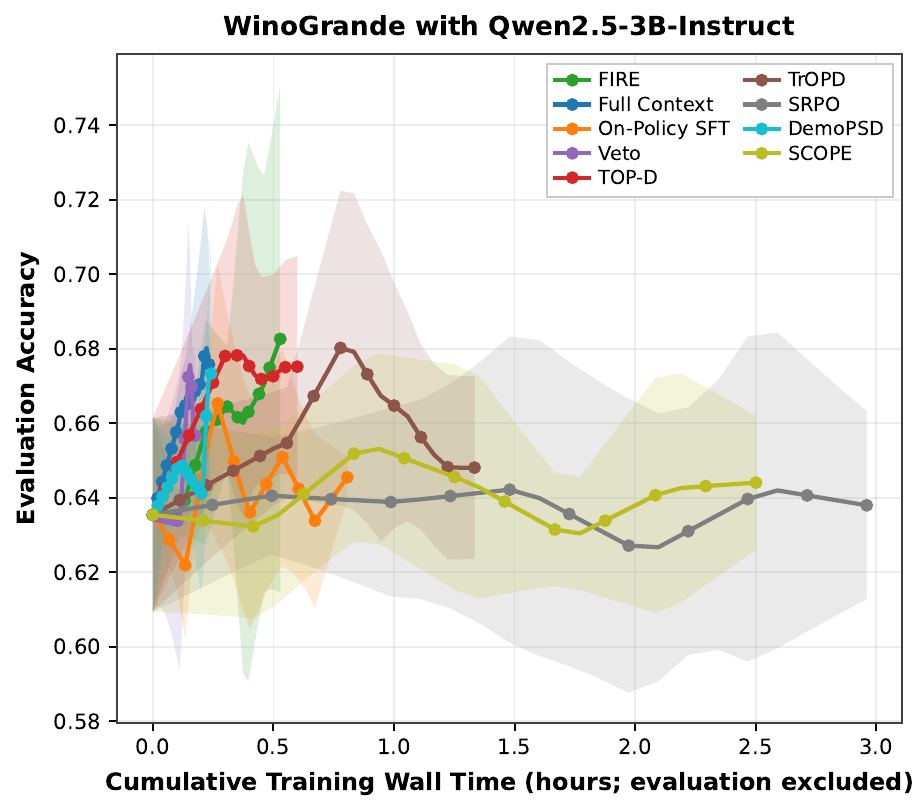}
    \caption*{WinoGrande}
\end{subfigure}
\hfill
\begin{subfigure}[t]{0.32\textwidth}
    \centering
    \includegraphics[width=\linewidth]{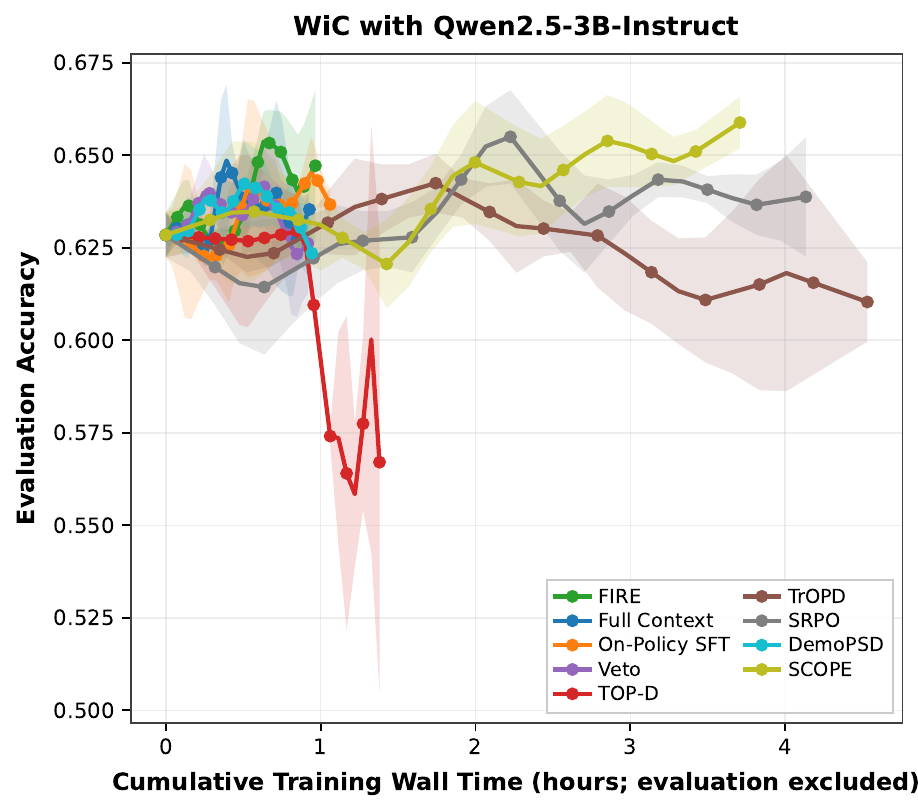}
    \caption*{WiC}
\end{subfigure}

\vspace{0.5em}

{\small\textbf{(b) Qwen2.5-3B-Instruct}}

\vspace{0.5em}

\caption{
Training time versus accuracy of each algorithm on (a) Qwen2.5-1.5B-Instruct and (b) Qwen2.5-3B-Instruct. FIRE offers the best tradeoff between training time and accuracy.
}
\label{appendix_exp:time_vs_acc}

\end{figure}

In Fig. \ref{appendix_exp:time_vs_acc}, we consider whether the observations in the previous subsection (Appendix \ref{appendix:tokens_gen_exp_section}) still hold in relation to the wall-clock training time. Similarly to Appendix \ref{appendix:tokens_gen_exp_section}, each run's trajectory is linearly interpolated onto a shared grid over the range jointly covered by all runs for that method. We can see from Fig. \ref{appendix_exp:time_vs_acc} that the observations still hold. FIRE offers the best tradeoff between training time and accuracy, with FIRE being multiple hours faster than some of the better performing baselines (e.g., SRPO). Similarly to Appendix \ref{appendix:tokens_gen_exp_section}, while we note SCOPE does better in final accuracy than FIRE on WiC, FIRE \textit{consistently performs well on all settings}, in comparison to Full Context, DemoPSD, SCOPE, and TOP-D, for example. Some of the baselines' slower performance can be attributed to requiring a smaller micro-batch size and therefore more gradient accumulation steps, as outlined in Appendix \ref{appendix:hyperparams}. However, this configuration was necessitated purely by resource constraints. Importantly, the wall-clock results exhibit characteristics similar to those in Fig. \ref{appendix_exp:gen_tokens_vs_acc}.  A consequence of these results is that despite the $\mathcal{O}(B)$ operational overhead for the leave-one-out counterfactuals in Sec. \ref{sec:fire}, FIRE still exhibits a better wall-clock time-accuracy tradeoff.

\subsection{Consideration for Large Model Size (14B)}\label{appendix_sec:14b}

In comparison to the smaller model sizes exhibited in Sec. \ref{sec:experiments}, we seek to see if FIRE's advantages will persist on large model sizes. Therefore, we consider Qwen2.5-14B-Instruct, fine-tuned on six categories of the more challenging MMLU-Pro \citep{wang2024mmlu} dataset. Specifically, we consider categories of biology, chemistry, computer science (CS), economics, engineering, and physics, which were picked via random selection. 

\subsubsection{Accuracy of Qwen2.5-14B-Instruct}

\begin{table*}[t]
\centering
\caption{Final accuracy (\%) on MMLU-Pro across subject domains using Qwen2.5-14B-Instruct. FIRE achieves the strongest overall performance across the evaluated methods.}
\label{appendix_tab:mmlu_pro_14b_results}
\resizebox{\textwidth}{!}{%
\begin{tabular}{llccccccc}
\toprule
\textbf{Model} & \textbf{Algorithm} & \textbf{Biology} & \textbf{Chemistry} &
\textbf{CS} & \textbf{Economics} & \textbf{Engineering} &
\textbf{Physics} & \textbf{Overall} \\
\midrule

\multirow{7}{*}{\textbf{Qwen2.5-14B-Instruct}}
& Full Context
& $75.00{\pm}2.95$
& $46.88{\pm}1.47$
& $54.17{\pm}2.95$
& $64.58{\pm}0.00$
& $40.62{\pm}1.47$
& $40.62{\pm}4.42$
& $53.65{\pm}0.25$ \\

& TOP-D
& $73.96{\pm}7.37$
& $46.88{\pm}4.42$
& $53.12{\pm}4.42$
& $66.67{\pm}0.00$
& $38.54{\pm}4.42$
& $42.71{\pm}7.37$
& $53.65{\pm}0.74$ \\

& TrOPD
& $68.75{\pm}0.00$
& $48.96{\pm}7.37$
& $53.12{\pm}4.42$
& $62.50{\pm}0.00$
& $40.62{\pm}7.37$
& $43.75{\pm}0.00$
& $52.95{\pm}1.72$ \\

& SRPO
& $73.96{\pm}1.47$
& $45.83{\pm}2.95$
& $52.08{\pm}2.95$
& $65.62{\pm}1.47$
& $40.62{\pm}4.42$
& $42.71{\pm}1.47$
& $53.47{\pm}0.98$ \\

& DemoPSD
& $77.08{\pm}2.95$
& $46.88{\pm}1.47$
& $54.17{\pm}2.95$
& $65.62{\pm}1.47$
& $39.58{\pm}0.00$
& $40.62{\pm}4.42$
& $53.99{\pm}1.23$ \\

& SCOPE
& $70.83{\pm}8.84$
& $20.83{\pm}0.00$
& $47.92{\pm}8.84$
& $55.21{\pm}7.37$
& $19.79{\pm}13.26$
& $27.08{\pm}5.89$
& $40.28{\pm}7.37$ \\

\rowcolor{blue!8}
& \textbf{FIRE (Ours)}
& $77.08{\pm}2.95$
& $52.08{\pm}2.95$
& $60.42{\pm}5.89$
& $62.50{\pm}0.00$
& $42.71{\pm}1.47$
& $38.54{\pm}1.47$
& $\mathbf{55.56}{\pm}1.47$ \\

\bottomrule
\end{tabular}%
}
\end{table*}

In Table \ref{appendix_tab:mmlu_pro_14b_results}, we present the overall performance of FIRE in comparison to the baselines on six categories of MMLU-Pro on Qwen2.5-14B-Instruct. From Table \ref{appendix_tab:mmlu_pro_14b_results} we can note that FIRE's performance advantages persist on a model that is significantly larger than the 1.5B and 3B sizes explored in Sec. \ref{sec:experiments}. Moreover, we can note that FIRE is the best performing method across four of the six explored categories, with high performance overall. Overall, these results indicate that FIRE's methodology is robust to differing model sizes, with performance not dependent on larger or smaller parameter sets.  

\subsubsection{Gradient Norm of Qwen2.5-14B-Instruct}

\begin{figure*}[t]
    \centering
    \includegraphics[width=1.0\linewidth]{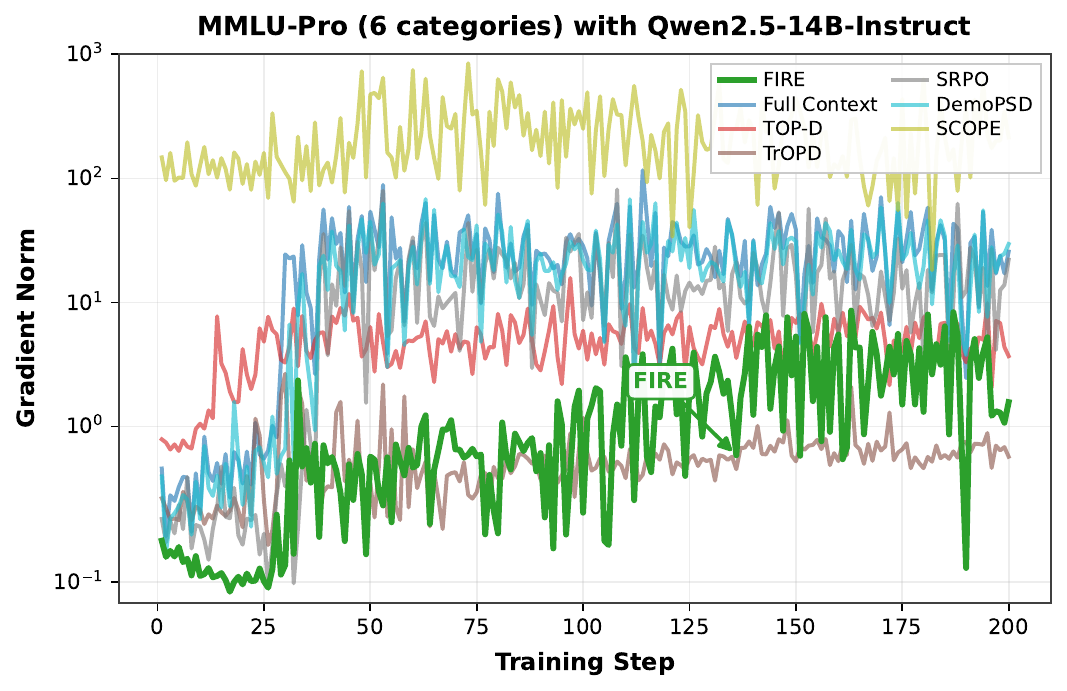}
    \caption{Gradient norms before clipping on six MMLU-Pro categories. Consistent with Sec. \ref{sec:experiments}, FIRE maintains controlled gradient magnitudes at the 14B scale relative to most baselines.}
    \label{appendix_fig:mmlu_pro_grad_norms}
\end{figure*}

Similar to Sec. \ref{sec:experiments}, we examine whether FIRE's optimization stability persists at a significantly larger model scale (Qwen2.5-14B-Instruct) in Fig. \ref{appendix_fig:mmlu_pro_grad_norms}. FIRE maintains substantially lower gradient norms than most baselines, with only TrOPD exhibiting consistently smaller magnitudes. While TOP-D shows somewhat lower oscillation, its gradient norms remain larger than FIRE's. Both TOP-D and TrOPD operate under a trust-region regime, which may explain their smoother gradient behavior. FIRE instead regulates updates through its Fisher-trace-based radius, yielding still relatively controlled gradients at the 14B scale. Although FIRE is not the smoothest method in this setting, the results remain consistent with the broader optimization stability observed in Sec. \ref{sec:experiments}. Moreover, TOP-D and TrOPD are not consistently smooth across all settings, in contrast with FIRE.

\subsection{Ablation Study on FIRE components}

\begin{table*}[t]
\centering
\caption{Final accuracy (\%) for FIRE component ablations across Qwen2.5-1.5B-Instruct and Qwen2.5-3B-Instruct and their associated datasets over three runs. Removing attribution, projection, or replacing projection with hard gradient excision degrades average performance, demonstrating the contribution of the full FIRE objective.}
\label{tab:fire_ablation}
\resizebox{\textwidth}{!}{%
\begin{tabular}{llcccc}
\toprule
\textbf{Model} & \textbf{Algorithm} & \textbf{GSM8K} & \textbf{ASDiv} &
\textbf{AQuA-RAT} & \textbf{Avg.} \\
\midrule

\multirow{4}{*}{\textbf{Qwen2.5-1.5B-Instruct}}
& No Attribution
                 & $61.11{\pm}1.20$ & $74.13{\pm}3.05$ & $50.35{\pm}4.21$ & 61.86 \\
& No Projection
                 & $56.77{\pm}7.22$ & $69.10{\pm}11.14$ & $51.04{\pm}0.52$ & 58.97 \\
& Hard Gradient Excision
                 & $58.33{\pm}4.13$ & $66.32{\pm}13.94$ & $53.12{\pm}1.80$ & 59.26 \\
\rowcolor{blue!8}
& \textbf{FIRE (Ours)}
                 & $60.94{\pm}2.71$ & $78.30{\pm}3.84$ & $52.26{\pm}2.46$
                 & \textbf{63.83} \\

\midrule
\textbf{Model} & \textbf{Algorithm} & \textbf{ARC-Challenge} &
\textbf{WinoGrande} & \textbf{WiC} & \textbf{Avg.} \\
\midrule

\multirow{4}{*}{\textbf{Qwen2.5-3B-Instruct}}
& No Attribution
                 & $85.94{\pm}2.27$ & $65.45{\pm}5.22$ & $65.28{\pm}0.80$ & 72.22 \\
& No Projection
                 & $86.63{\pm}2.17$ & $66.32{\pm}6.04$ & $64.06{\pm}2.76$ & 72.34 \\
& Hard Gradient Excision
                 & $86.28{\pm}2.57$ & $66.49{\pm}5.00$ & $64.41{\pm}1.20$ & 72.40 \\
\rowcolor{blue!8}
& \textbf{FIRE (Ours)}
                 & $86.11{\pm}2.46$ & $68.23{\pm}6.83$ & $64.76{\pm}2.10$
                 & \textbf{73.03} \\

\bottomrule
\end{tabular}%
}
\end{table*}

Here, we conduct ablation studies on the individual components of the FIRE methodology. For No Attribution, we consider the exclusion of leave-one-out counterfactuals and utilize the full feedback (i.e., no \eqref{eq:fire_attribution_target}). For No Projection, the final radial projection for incorrect responses is not conducted (i.e., no \eqref{eq:fire_contraction_coefficient} and \eqref{eq:fire_final_target}). Hard Gradient Excision corresponds to removing the feedback field with the largest leave-one-out influence on the teacher-induced gradient and distilling on the resulting counterfactual target. For the Qwen2.5-1.5B-Instruct, we use a uniformly higher learning-rate setting across all variants to better emphasize component behavior advantages of FIRE in more unstable optimization landscapes. For AQuA-RAT, the peak $a_k = 5 \times 10^{-6}$ with $a_0 = 5 \times 10^{-7}$ over 200 steps. For GSM8K and ASDiv, it is $6 \times 10^{-6}$/$1 \times 10^{-6}$ and $1 \times 10^{-5}$/$1 \times 10^{-6}$ respectively. 

From the results in Table \ref{tab:fire_ablation}, we note FIRE's performance advantages. In particular, we observe that FIRE outperforms all three ablations across both Qwen2.5-1.5B-Instruct and Qwen2.5-3B-Instruct on average. Therefore, the individual components that are introduced in the FIRE methodology result in an overall net positive improvement of performance. 

\subsection{Learning Rate Impact}
Since the nominal rate $a_0$ hyperparameter is designed to cope with potential instability in higher learning rate regimes, we examine whether $a_0$ properly plays the role of stabilizing training across differing learning rates. For this, we consider $a_0$ set to $1.0 \times 10^{-7}$, $5.0 \times 10^{-7}$, $1.0 \times 10^{-8}$, $5.0 \times 10^{-7}$, $5.0 \times 10^{-7}$, and $5.0 \times 10^{-8}$ for GSM8K, ASDiv, AQuA-RAT, ARC-Challenge, WinoGrande, and WiC respectively. 

\subsubsection{Ablation on Differing Learning Rates with Adjusted Nominal Rate}\label{appendix_subsec:lr_table}

\begin{table}[htbp]
\centering
\caption{Learning-rate ablation for FIRE. We sweep the learning rate over
$\{0.5\times, 1\times, 2\times, 4\times\}$ the default learning rate utilized for each dataset-model pair.}
\label{appendix_tab:fire_lr_ablation}
\begin{tabular}{lccc}
\toprule
\multicolumn{4}{c}{\textbf{Qwen2.5-1.5B-Instruct}} \\
\midrule
\textbf{LR Multiplier} & \textbf{GSM8K} & \textbf{ASDiv} & \textbf{AQuA-RAT} \\
\midrule
$0.5\times$
& $54.51{\pm}5.22$ & $76.91{\pm}4.55$ & $54.86{\pm}1.59$ \\
$1\times$
& $63.72{\pm}3.01$ & $81.94{\pm}3.35$ & $53.82{\pm}1.59$ \\
$2\times$
& $61.63{\pm}4.21$ & $81.77{\pm}2.39$ & $52.78{\pm}2.41$ \\
$4\times$
& $54.34{\pm}4.21$ & $78.47{\pm}1.67$ & $50.17{\pm}1.31$ \\
\midrule
\multicolumn{4}{l}{\footnotesize LR ($1\times$): GSM8K/ASDiv $=4.0\times10^{-6}$; AQuA-RAT $=2.0\times10^{-6}$.} \\
\bottomrule
\end{tabular}

\vspace{0.8em}

\begin{tabular}{lccc}
\toprule
\multicolumn{4}{c}{\textbf{Qwen2.5-3B-Instruct}} \\
\midrule
\textbf{LR Multiplier} & \textbf{ARC-Challenge} & \textbf{WinoGrande} & \textbf{WiC} \\
\midrule
$0.5\times$
& $83.85{\pm}2.39$ & $65.80{\pm}0.30$ & $59.72{\pm}5.35$ \\
$1\times$
& $83.16{\pm}1.08$ & $67.53{\pm}0.60$ & $61.46{\pm}4.51$ \\
$2\times$
& $82.64{\pm}1.59$ & $66.84{\pm}2.41$ & $58.51{\pm}5.76$ \\
$4\times$
& $81.77{\pm}1.88$ & $65.28{\pm}3.01$ & $55.56{\pm}6.01$ \\
\midrule
\multicolumn{4}{l}{\footnotesize LR ($1\times$): ARC-Challenge/WinoGrande $=4.0\times10^{-6}$; WiC $=1.0\times10^{-6}$.} \\
\bottomrule
\end{tabular}
\end{table}

In Table \ref{appendix_tab:fire_lr_ablation}, we sweep over four differing learning rates across Qwen2.5-1.5B-Instruct and Qwen2.5-3B-Instruct. While we note that setting the learning rate too high can still lead to suboptimal performance, we observe that FIRE successfully prevents collapse across all observed learning rates and still maintains acceptable levels of performance in all settings.

\subsubsection{Gradient Norm Stability Across Learning Rates}

\begin{figure}[t]
\centering

\begin{subfigure}[t]{0.32\textwidth}
    \centering
    \includegraphics[width=\linewidth]{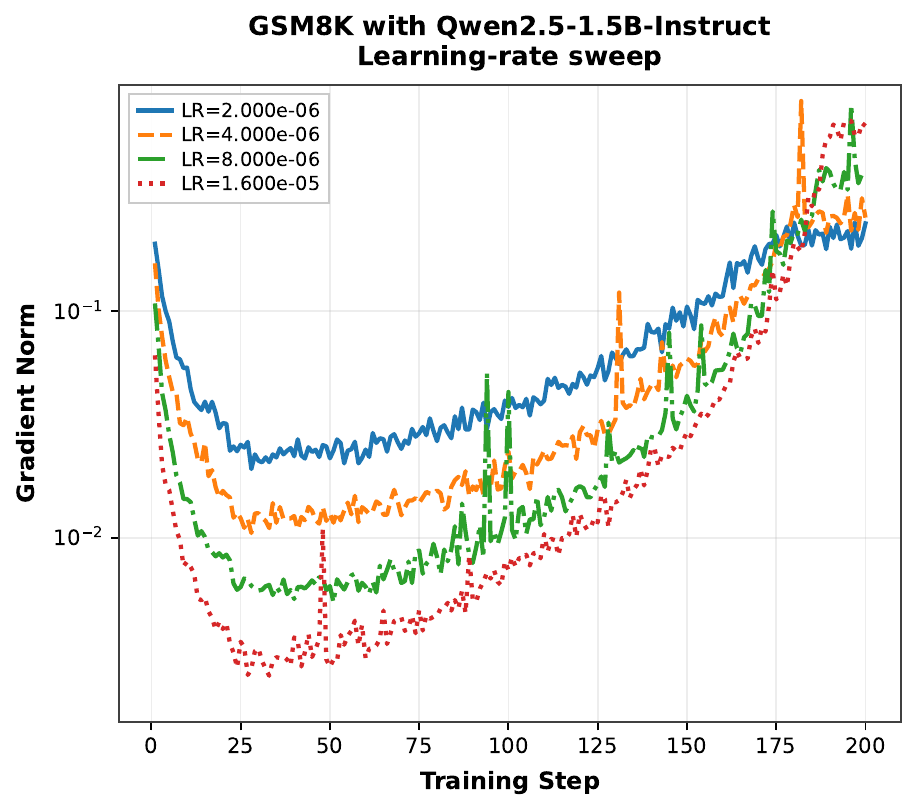}
    \caption*{GSM8K}
\end{subfigure}
\hfill
\begin{subfigure}[t]{0.32\textwidth}
    \centering
    \includegraphics[width=\linewidth]{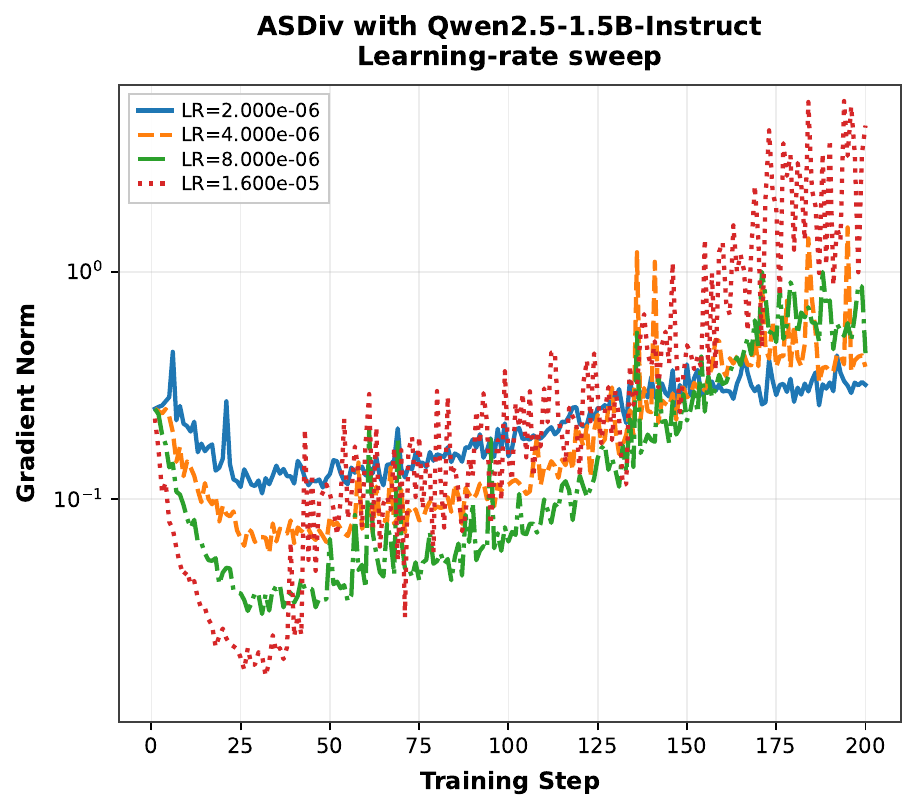}
    \caption*{ASDiv}
\end{subfigure}
\hfill
\begin{subfigure}[t]{0.32\textwidth}
    \centering
    \includegraphics[width=\linewidth]{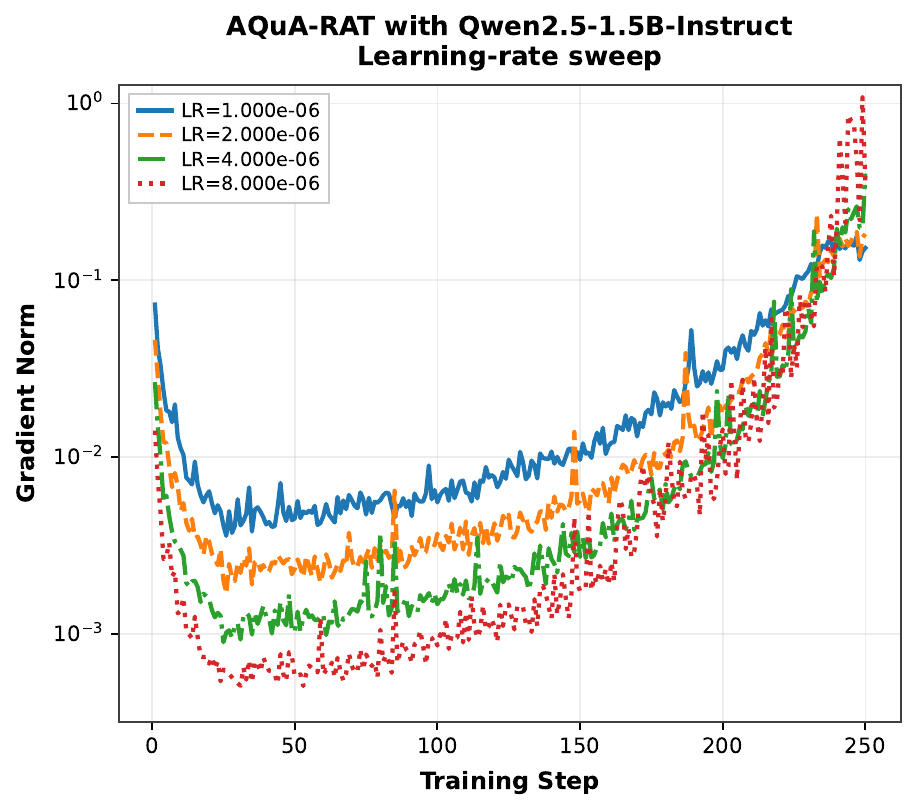}
    \caption*{AQuA-RAT}
\end{subfigure}

\vspace{0.5em}

{\small\textbf{(a) Qwen2.5-1.5B-Instruct}}

\vspace{0.75em}

\begin{subfigure}[t]{0.32\textwidth}
    \centering
    \includegraphics[width=\linewidth]{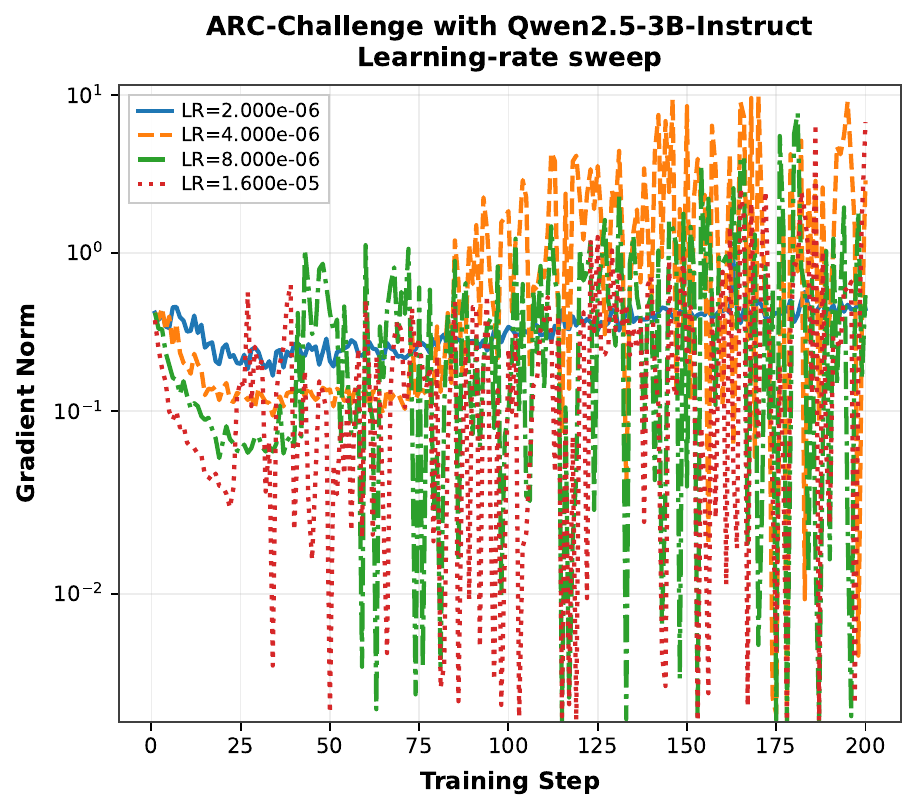}
    \caption*{ARC-Challenge}
\end{subfigure}
\hfill
\begin{subfigure}[t]{0.32\textwidth}
    \centering
    \includegraphics[width=\linewidth]{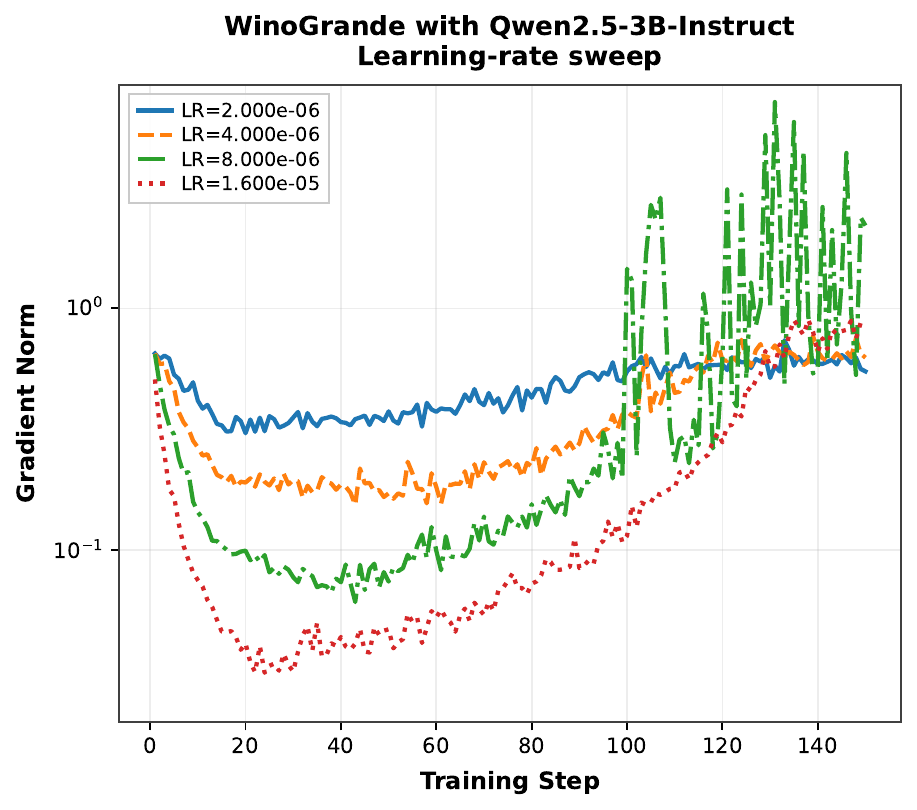}
    \caption*{WinoGrande}
\end{subfigure}
\hfill
\begin{subfigure}[t]{0.32\textwidth}
    \centering
    \includegraphics[width=\linewidth]{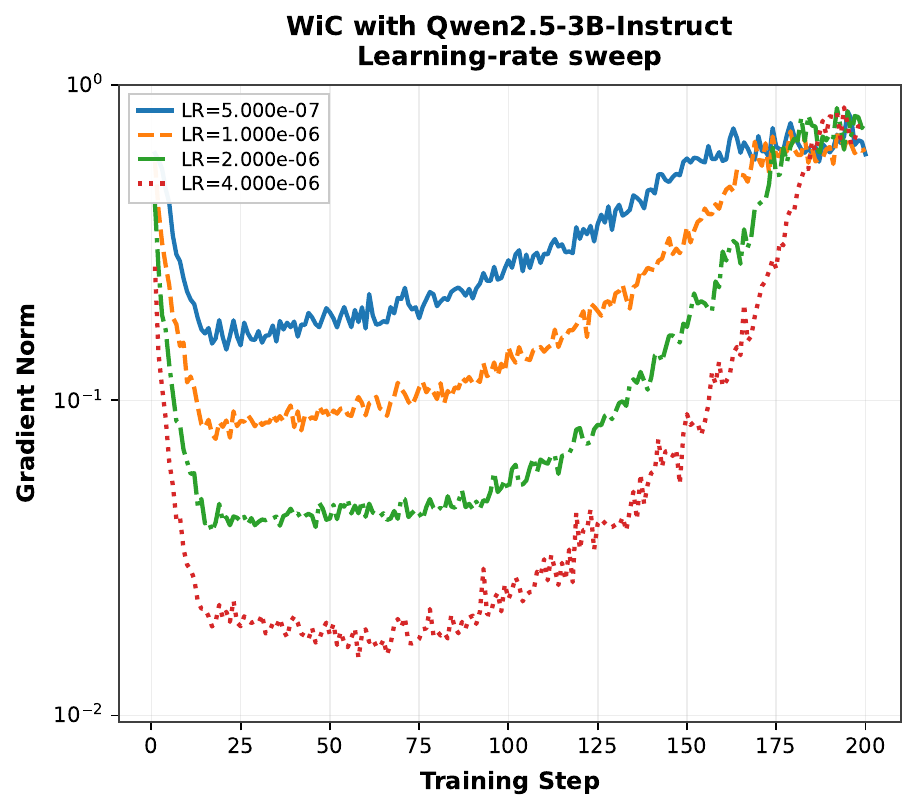}
    \caption*{WiC}
\end{subfigure}

\vspace{0.5em}

{\small\textbf{(b) Qwen2.5-3B-Instruct}}

\vspace{0.5em}

\caption{
Gradient norms on (a) Qwen2.5-1.5B-Instruct and (b) Qwen2.5-3B-Instruct across varying learning rates.
}
\label{appendix_exp:lr_grad_norms}

\end{figure}

In addition, we consider how the varying learning rates affect the norms of the gradients in Fig. \ref{appendix_exp:lr_grad_norms}. We note across all datasets with the exception of ARC-Challenge, the gradient norms stay relatively stable across both lower and larger learning rate regimes, with most settings below the clipping threshold of $1.0$. This indicates that the observations in Appendix \ref{appendix_subsec:lr_table} correspond mostly with the observations in Fig. \ref{appendix_exp:lr_grad_norms}. While we note relatively high oscillation for ARC-Challenge across three differing learning rates, we would emphasize the relative magnitude of the gradients \textit{are small in comparison to the baselines}. For example, with ARC-Challenge, the magnitude never peaks past $10.0$, while they nearly reach $100.0$ on DemoPSD, Full Context, Veto, and TOP-D, with the gradient norms exceeding $100.0$ for SCOPE in Fig. \ref{fig:main_grad_norms}. Therefore, even in the higher learning rate scenario, FIRE achieves substantially more constrained gradients than multiple baselines under a more stable learning rate regime.  


\section{Pseudocode of FIRE}\label{appendix:pseudocode}

Following the notation of Sec.~\ref{sec:fire}, Algorithm~\ref{alg:fire} gives the
optimizer-step implementation of FIRE. All quantities used to construct
$\beta^t$, $\alpha_j^t$, $\eta^t$, and the teacher targets are detached. Gradients are taken only through the live student distribution in the final branch loss.

\begin{algorithm}[H]
\caption{FIRE at optimizer step $k$}
\label{alg:fire}
\footnotesize
\begin{algorithmic}[1]
\Require Student $\pi_{\boldsymbol{\theta}_k}$, EMA teacher
$\pi_{\bar{\boldsymbol{\theta}}_k}$, minibatch $\mathcal{B}$, verifier $\omega$,
feedback builder $\mathcal{F}$, learning rate $a_k$, nominal rate $a_0$
\ForAll{$\mathbf{x}_i\in\mathcal{B}$}
    \State Sample one on-policy response
    $\mathbf{y}_i\sim\pi_{\boldsymbol{\theta}_k}(\cdot\mid\mathbf{x}_i)$.
    \State Score the fixed response with the live student, set
    $p_i^t\gets\operatorname{sg}[p_{\boldsymbol{\theta}_k,i}^t]$ and compute
    $\rho_{k,i}^t$ by \eqref{eq:fire_radius}
    \If{$\omega(\mathbf{y}_i)=1$} \Comment{Correct branch}
        \State Compute
        $\mathbf{g}_{+,i}^t=p_i^t-\mathbf{e}_{y_{i,t}}$ and
        $\beta_i^t=\min(1,\rho_{k,i}^t/\lVert\mathbf{g}_{+,i}^t\rVert_2)$
        \State $\ell_i\gets$ token-averaged re-weighted on-policy SFT from \eqref{eq:fire_correct}
    \Else \Comment{Incorrect branch}
        \State Build
        $\mathbf{c}_i=\mathcal{F}(\mathbf{x}_i,\mathbf{y}_i)
        =(b_{i,1},\ldots,b_{i,B})$
        \State Teacher-force the EMA on the same fixed
        $\mathbf{y}_i$ under $\mathbf{c}_i$ and each
        $\mathbf{c}_{i,\setminus j}$ to obtain
        $q_{\mathbf{c},i}^t$ and $\{q_{-j,i}^t\}_{j=1}^{B}$
        \State Compute gradients and energies
        $\mathbf{g}^t(\cdot)$, $E_{\mathbf{c},i}^t$, and
        $\{E_{j,i}^t\}_{j=1}^{B}$ using
        \eqref{eq:induced_gradient} and~\eqref{eq:fire_gradient_energies}
        \State Compute $\chi_i^t$, $\{w_{j,i}^t\}_{j=1}^{B}$, and
        $\{\alpha_{j,i}^t\}_{j=0}^{B}$ using
        \eqref{eq:fire_evidence}--\eqref{eq:fire_attribution_weights}
        \State Construct $q_{A,i}^t$, then $\eta_i^t$ and
        $q_{\mathrm{FIRE},i}^t$ using
        \eqref{eq:fire_attribution_target}--\eqref{eq:fire_final_target}
        \State $\ell_i\gets M_i^{-1}\sum_t m_{i,t}
        \operatorname{KL}\!\left(
        p_{\boldsymbol{\theta},i}^t\middle\|
        \operatorname{sg}[q_{\mathrm{FIRE},i}^t]\right)$
    \EndIf
\EndFor
\State $\mathcal{L}_{\mathrm{FIRE}}\gets
|\mathcal{B}|^{-1}\sum_{i\in\mathcal{B}}\ell_i$
\State $\boldsymbol{\theta}_{k+1}\gets
\operatorname{OptStep}\!\left(
\boldsymbol{\theta}_k,
\nabla_{\boldsymbol{\theta}_k}\mathcal{L}_{\mathrm{FIRE}},a_k\right)$
\State $\bar{\boldsymbol{\theta}}_{k+1}\gets
\mu\bar{\boldsymbol{\theta}}_k+(1-\mu)\boldsymbol{\theta}_{k+1}$
\end{algorithmic}
\end{algorithm}


\section{Hyperparameter Settings}\label{appendix:hyperparams}

This section reports the settings used for the primary experiments in Sec. \ref{sec:experiments} and
scaling experiments. The optimizer learning rate is shared by FIRE and the corresponding baselines for a given model-dataset pair. FIRE's nominal rate $a_0$ affects only
the FIRE radius in~\eqref{eq:fire_radius}.

Across experiments, we use AdamW with
$\beta_1=0.9$, $\beta_2=0.95$, weight decay $0.01$, linear warmup followed by
cosine decay, and gradient clipping at norm $1.0$. We use \texttt{bfloat16}, gradient checkpointing, and LoRA with rank $16$, LoRA scaling $32$, and dropout
$0.0$ on the \texttt{q\_proj}, \texttt{k\_proj}, \texttt{v\_proj},
\texttt{o\_proj}, \texttt{gate\_proj}, \texttt{up\_proj}, and
\texttt{down\_proj} modules. The effective batch size is $8$ in all resolved runs. Generation uses top-$p=0.95$ and temperature $0.8$, except On-Policy SFT, which uses temperature $1.0$. On-Policy SFT uses a differing sampling temperature because its sampled trajectories are directly reused as hard supervised targets. All distillation-based methods use the same sampling temperature. The maximum prompt and feedback budgets are
$1280$ and $768$ tokens, respectively. FIRE uses four feedback fields during implementation.

\subsection{Model and Dataset-Specific Settings}

\begin{table}[H]
\centering
\caption{Resolved model- and dataset-specific settings. The EMA column reports
the student update weight $1-\mu$ in \eqref{eq:sdpo_updates}. The nominal
rate $a_0$ is used only by FIRE.}
\label{tab:main_run_hparams}
\resizebox{\textwidth}{!}{%
\begin{tabular}{llccccccccc}
\toprule
\textbf{Model} & \textbf{Dataset} & \textbf{Peak LR} & \textbf{$a_0$}
& \textbf{$1-\mu$} & \textbf{Warmup}
& \textbf{Steps} & \textbf{Max context} & \textbf{Max new}
& \textbf{Train pool} & \textbf{Eval. $n$} \\
\midrule
Qwen2.5-1.5B & GSM8K         & $4{\times}10^{-6}$ & $4{\times}10^{-6}$ & 0.03 & 25 & 200 & 2560 & 384 & 1024 & 192 \\
Qwen2.5-1.5B & ASDiv         & $4{\times}10^{-6}$ & $4{\times}10^{-6}$ & 0.03 & 25 & 200 & 2304 & 320 & 1600 & 192 \\
Qwen2.5-1.5B & AQuA-RAT      & $2{\times}10^{-6}$ & $2{\times}10^{-6}$ & 0.03 & 25 & 250 & 3072 & 512 & 2048 & 192 \\
\midrule
Qwen2.5-3B   & ARC-Challenge & $4{\times}10^{-6}$ & $4{\times}10^{-6}$ & 0.01 & 20 & 200 & 2560 & 320 & 2048 & 192 \\
Qwen2.5-3B   & WinoGrande    & $4{\times}10^{-6}$ & $4{\times}10^{-6}$ & 0.01 & 15 & 150 & 2560 & 320 & 2048 & 192 \\
Qwen2.5-3B   & WiC           & $1{\times}10^{-6}$ & $1{\times}10^{-6}$ & 0.01 & 15 & 200 & 2560 & 320 & 2048 & 192 \\
\midrule
Qwen2.5-14B  & MMLU-Pro (six categories) & $1{\times}10^{-5}$ & $2{\times}10^{-6}$ & 0.01 & 20 & 200 & 3072 & 256 & 1536 & 288 \\
\bottomrule
\end{tabular}%
}
\end{table}

\subsection{Method-Specific Settings}

Table~\ref{tab:method_specific_hparams} lists only the principal
method-specific hyperparameters. Pure implementation constants such as numerical epsilons are excluded. We retain the notation of the original baseline papers for baseline-specific hyperparameters. Some symbols may therefore overlap with notation used elsewhere in this paper, but such symbols are constrained to only Table~\ref{tab:method_specific_hparams} and should not be interpreted as referring to the quantities defined in the main text.

\begin{table}[H]
\centering
\caption{Important method-specific hyperparameters for run
configurations. ``Small'' denotes the Qwen2.5-1.5B/3B experiments, and
``scaling'' denotes the larger model experiments.}
\label{tab:method_specific_hparams}
\resizebox{\textwidth}{!}{%
\begin{tabular}{lp{0.78\textwidth}}
\toprule
\textbf{Method} & \textbf{Method-specific settings} \\
\midrule
Full Context &
No additional method-specific hyperparameter beyond the shared optimizer,
generation, EMA, and LoRA settings. \\
On-Policy SFT &
Group size $4$ (1.5B) / $2$ (3B). \\
Veto &
$\beta$ decays linearly from $0.8$ to $0.0$, forward-KL objective. \\
TOP-D &
Group size $4$, proximal coefficient $\alpha=0.2$, PPO clipping
$\epsilon=0.2$, one off-policy epoch, one prompt per internal minibatch. \\
TrOPD &
Group size $4$, guidance coefficient
$\beta=10^{-3}$, maximum guided prefix length $384$. \\
SRPO &
Group size $4$ for the small-model experiments and $2$ for scaling, entropy coefficient $\beta=1.0$, PPO clipping $\epsilon=0.2$, ratio cap $2.0$. \\
DemoPSD &
Maximum leakage coefficient $\alpha_{\max}=0.15$, disagreement scale $\beta=50$. \\
SCOPE &
Group size $4$ for the small-model experiments and $2$ for scaling, group-weight temperature $1.0$. \\
FIRE (ours) &
Four structured feedback fields, nominal rate $a_0$ as in
Table~\ref{tab:main_run_hparams}, excess-$\ell_2$ gradient-energy attribution, learning-rate step normalization and Fisher-informed re-weighted correct-branch SFT enabled. \\
\bottomrule
\end{tabular}%
}
\end{table}

\subsection{Micro-Batch Sizes and Gradient Accumulation}

\begin{table}[H]
\centering
\caption{Resolved micro-batch size / gradient-accumulation steps. All entries
have effective batch size $8$. A dash denotes a method not used in that reported
model-scale comparison.}
\label{tab:batching_hparams}
\begin{tabular}{lccc}
\toprule
\textbf{Method} & \textbf{Qwen 1.5B} & \textbf{Qwen 3B}
& \textbf{Qwen 14B} \\
\midrule
Full Context    & $8/1$ & $8/1$ & $2/4$ \\
On-Policy SFT   & $2/4$ & $2/4$ & --   \\
Veto            & $8/1$ & $8/1$ & -- \\
TOP-D           & $2/4$ & $2/4$ & $1/8$    \\
TrOPD           & $2/4$ & $2/4$ & $1/8$\\
SRPO            & $2/4$ & $2/4$ & $1/8$ \\
DemoPSD         & $8/1$ & $8/1$ & $2/4$ \\
SCOPE           & $2/4$ & $2/4$ & $1/8$ \\
FIRE (ours)     & $8/1$ & $8/1$ & $2/4$ \\
\bottomrule
\end{tabular}
\\[-0.1em]
\end{table}

Table \ref{tab:batching_hparams} reports the micro-batch size and number of gradient accumulation steps used for each method while maintaining the same effective batch size. For baselines requiring more accumulation steps than FIRE or Full Context, this choice was necessitated solely by resource constraints, rather than to confer a computational advantage on FIRE or Full Context. Moreover, Appendix \ref{appendix:tokens_gen_exp_section} and Fig. \ref{appendix_exp:gen_tokens_vs_acc} evaluate accuracy as a function of the number of autoregressively generated tokens, demonstrating that FIRE achieves a more favorable accuracy-generation trade-off. This provides evidence that FIRE's computational efficiency gains are intrinsic to the method rather than due to the batching configuration.


\section{Prompt and Feedback Templates}
\label{appendix:templates}

\subsection{Per-dataset Templates}
We provide the prompt templates used in our experiments below. Text in
\textcolor{promptblue}{\{blue braces\}} denotes fields populated from the
corresponding dataset or sampled trajectory.  For FIRE, the teacher feedback context shown below is used for incorrect responses. Correct responses are instead
routed to Fisher-informed re-weighted on-policy SFT as described in Sec.~\ref{sec:method}.


\begin{prompttemplate}{GSM8K}
\ttfamily\small

Solve the following task.\par\medskip

Work through the problem step by step, showing your arithmetic.
Then, on a new final line, write the final answer as
\texttt{\#\#\#\# } followed by the number only
(for example \texttt{\#\#\#\# 42}).
Write nothing after that line.\par\medskip

\pvar{question}

\teacherfeedback
\end{prompttemplate}


\begin{prompttemplate}{ASDiv}
\ttfamily\small

Solve the following task.\par\medskip

Work through the problem step by step, showing your arithmetic.
Then, on a new final line, write the final answer as
\texttt{\#\#\#\# } followed by the number only
(for example \texttt{\#\#\#\# 42}).
Write nothing after that line.\par\medskip

\pvar{body}\par

\pvar{question}

\teacherfeedback
\end{prompttemplate}


\begin{prompttemplate}{AQuA-RAT}
\ttfamily\small

Solve the following task.\par\medskip

Reason step by step about the options. Then, on a new final line,
write \texttt{Answer: } followed by the single option letter only
(for example \texttt{Answer: B}). Write nothing after that line.\par\medskip

\pvar{question}\par\medskip

Choices:\par
A. \pvar{choice A}\par
B. \pvar{choice B}\par
C. \pvar{choice C}\par
D. \pvar{choice D}\par
E. \pvar{choice E}

\teacherfeedback
\end{prompttemplate}


\begin{prompttemplate}{ARC-Challenge}
\ttfamily\small

Solve the following task.\par\medskip

Reason step by step about the options. Then, on a new final line,
write \texttt{Answer: } followed by the single option letter only
(for example \texttt{Answer: B}). Write nothing after that line.\par\medskip

\pvar{question}\par\medskip

Choices:\par
\pvar{normalized lettered choices}

\teacherfeedback
\end{prompttemplate}


\begin{prompttemplate}{WinoGrande}
\ttfamily\small

Solve the following task.\par\medskip

Reason step by step about the options. Then, on a new final line,
write \texttt{Answer: } followed by the single option letter only
(for example \texttt{Answer: B}). Write nothing after that line.\par\medskip

Choose the option that best fills the blank in the sentence.
Resolve the reference using commonsense reasoning.\par\medskip

Sentence:\par
\pvar{sentence}\par\medskip

Choices:\par
A. \pvar{option 1}\par
B. \pvar{option 2}

\teacherfeedback
\end{prompttemplate}


\begin{prompttemplate}{WiC}
\ttfamily\small

Solve the following task.\par\medskip

Reason step by step about the options. Then, on a new final line,
write \texttt{Answer: } followed by the single option letter only
(for example \texttt{Answer: B}). Write nothing after that line.\par\medskip

Consider the word '\pvar{word}' in the two sentences below.
Determine whether it has the same meaning in both sentences.\par\medskip

Sentence 1:\par
\pvar{sentence 1}\par\medskip

Sentence 2:\par
\pvar{sentence 2}\par\medskip

Choices:\par
A. Different meaning\par
B. Same meaning

\teacherfeedback
\end{prompttemplate}


\begin{prompttemplate}{MMLU-Pro}
\ttfamily\small

Solve the following task.\par\medskip

Reason step by step about the options. Then, on a new final line,
write \texttt{Answer: } followed by the single option letter only
(for example \texttt{Answer: B}). Write nothing after that line.\par\medskip

\pvar{question}\par\medskip

Choices:\par
\pvar{lettered non-N/A choices}

\teacherfeedback
\end{prompttemplate}

\subsection{Feedback Field Construction}
\label{appendix:feedback_fields}

For feedback-conditioned teacher supervision, we represent the feedback context as four structured fields, $\mathbf{c}=(b_1,b_2,b_3,b_4)$. Each field captures a distinct aspect of the
feedback signal, allowing FIRE to measure its contribution independently through the leave-one-field-out targets in~\eqref{eq:loo_teacher}.

\paragraph{Verifier result ($b_1$).}
The verifier field provides the primary outcome-level feedback. It records
whether the sampled response is correct, the student's normalized submitted
answer, and the accepted or expected final answer. For an incorrect response,
it additionally specifies the corrected final-answer line and instructs the
teacher to revise the final answer while retaining the task-specific output
format. Therefore, this field contains the most direct information about
\emph{what} in the final prediction must change.

\paragraph{Parser record ($b_2$).}
The parser field describes how the student's final prediction was extracted
from its sampled response. It records the task and expected answer types,
required final-answer marker, number of occurrences of that marker, whether
parsing succeeded, the extracted normalized answer, and a snapshot of the
final response line. For numerical tasks, it also records nearby numerical
tokens. This field therefore exposes whether an apparent error may be related
to answer extraction or the structure of the submitted response.

\paragraph{Verifier provenance ($b_3$).}
The provenance field supplies task and verifier-level context used to
interpret the verification decision. It includes the dataset adapter and task
type, a task fingerprint, available choice labels or numerical cues from the
prompt, the checker mode (e.g., exact matching or symbolic equivalence), and
the answer-normalization procedure. It additionally summarizes basic response
statistics such as character, nonempty-line, and numerical-token counts.
This field provides contextual evidence about \emph{how} the correctness
decision was produced without changing the verifier outcome itself.

\paragraph{Response-format diagnostics ($b_4$).}
The final field captures compliance with the required response format. It
records the expected final-answer marker and parsed result, identifies issues
such as a missing marker, repeated markers, or a non-canonical final line,
and indicates whether reasoning precedes the final answer. For arithmetic
responses, it also summarizes detected arithmetic expressions. The field
concludes with the task-level requirement that the response terminate with
exactly one valid final-answer line and no subsequent text.

\paragraph{Illustrative feedback instance.}
To make the structured feedback representation more understandable, the following box shows an illustrative example for an incorrect mathematical response on GSM8K. 

\begin{tcolorbox}[
  enhanced,
  breakable,
  colback=promptbg,
  colframe=promptpink,
  boxrule=0.8pt,
  arc=2.5mm,
  left=4mm,
  right=4mm,
  top=5mm,
  bottom=4mm,
  title={Example Structured Feedback Context},
  fonttitle=\bfseries\large\sffamily,
  coltitle=white,
  boxed title style={
    colback=promptpink,
    colframe=promptpink,
    boxrule=0pt,
    arc=0pt
  },
  attach boxed title to top left={
    xshift=5mm,
    yshift=-2.5mm
  },
  before skip=8pt,
  after skip=10pt
]

\ttfamily\small

\textbf{[Context block 1: Verifier result]}\par
Verifier feedback: source=reference\_checker;
status=incorrect; decision=final;
expected/accepted final answer=`42';
submitted normalized answer=`40';
replace the submitted final answer with `42'.\par

\medskip
\hrule
\medskip

\textbf{[Context block 2: Parser record]}\par
Parser diagnostics: source=response\_parser;
task uid=`example\_gsm8k\_001';
task type=math;
expected answer type=numeric;
final marker=`\#\#\#\#';
marker count=1;
parse success=yes;
extracted normalized final answer=`40';
final line snapshot=`\#\#\#\# 40';
numeric tokens near parsed span=[40].\par

\medskip
\hrule
\medskip

\textbf{[Context block 3: Verifier provenance]}\par
Context provenance: source=environment\_audit;
dataset adapter=gsm8k;
task type=math;
task fingerprint=`example\_001';
prompt numeric cues=[6, 7];
checker mode=numeric-equivalence;
normalization=task\_adapter\_final\_answer;
submitted normalized answer=`40';
response chars=184;
response nonempty lines=5;
response numeric-token count=8.\par

\medskip
\hrule
\medskip

\textbf{[Context block 4: Response-format diagnostics]}\par
Response-format diagnostics: source=format\_checker;
required final marker=`\#\#\#\#';
final-line parse result=`40';
format issues=none;
reasoning text before final line=yes;
arithmetic expression count=3;
arithmetic expression snippets=`6 $\times$ 7 = 40';
instruction=end with exactly one task-normal final-answer line
and no text after it.

\end{tcolorbox}

\section{Comparison with Closely Related Methods}\label{appendix:novelty_info}

Several recent methods address instability in on-policy distillation, but differ from FIRE in both setting and mechanism. In particular, some methods below study standard teacher-student on-policy distillation, whereas FIRE targets feedback-based \emph{self-distillation}, where the teacher and student are derived from the same model. In this section, we clarify in detail how FIRE differs from existing works.

\paragraph{SRPO and SCOPE.}
SRPO~\citep{li2026unifying} and SCOPE~\citep{zheng2026scope} also route correct
and incorrect trajectories differently. SRPO operates in a self-distillation setting, sending correct samples to GRPO and failed samples to self-distillation. SCOPE instead uses conventional teacher-student on-policy distillation, with student-perplexity-weighted on-policy SFT for correct trajectories and teacher-perplexity-weighted distillation for incorrect ones. FIRE, by contrast, uses Fisher-informed re-weighted on-policy SFT for correct responses and, for incorrect responses, attributes the induced gradient to individual feedback fields. Neither method performs individual feedback-field attribution or constrains both branches with Fisher and gradient-derived insights.

\paragraph{DemoPSD.}
DemoPSD~\citep{li2026demopsd} is also an on-policy self-distillation method and is the closest baseline in terms of target recalibration. It measures teacher-student disagreement using per-token Jensen-Shannon Divergence (JSD) and geometrically shifts the privileged teacher toward the student when disagreement is large. FIRE, by contrast, measures the gradient induced by the feedback-conditioned teacher. The full teacher is unchanged when its gradient lies within the local budget, with only unsafe updates triggering feedback-field attribution and, when still necessary, subsequent contraction. Therefore, consideration for individual feedback field influence via leave-one-out counterfactuals is not considered with DemoPSD. 

\paragraph{Veto.}
Veto~\citep{jang2026stable} studies conventional teacher-student on-policy distillation rather than self-distillation. It constructs a geometric teacher-student target with a scheduled interpolation coefficient. FIRE is instead token-specific and determined by the attribution-aware gradient relative to a Fisher-derived radius. FIRE also routes by correctness and attributes unsafe updates to individual feedback fields.

\paragraph{TOP-D and TrOPD.}
TOP-D~\citep{xie2026trust} and TrOPD~\citep{xing2026trust} likewise address conventional on-policy distillation with distinct teacher and student models, rather than the feedback-based self-distillation setting. While TOP-D and TrOPD also constrain unreliable or overly large distillation updates, their trust-region mechanisms differ from FIRE's Fisher-informed local gradient radius. TOP-D constructs a proximal teacher and uses internal trust-region iterations, while TrOPD identifies teacher-verifiable regions from teacher-student agreement and treats outliers separately. FIRE instead directly
constrains the token-level gradient induced by a feedback-conditioned self-teacher using the student's local Fisher geometry. In addition, neither TOP-D nor TrOPD routes trajectories by correctness into separate correct and incorrect-response objectives.

\paragraph{Key distinction.}
FIRE combines outcome routing, feedback-field counterfactual attribution in gradient space, and a token-level Fisher-informed update bound. The attribution determines \emph{which feedback-dependent direction} to follow, while the Fisher radius determines
\emph{how far} the update may move. Importantly, FIRE leaves the original full-feedback target unchanged whenever its induced update is already within this radius and outperforms all the aforementioned baselines in Sec. \ref{sec:experiments}.

\section{Additional FIRE Details}
\label{app:fire_details}

This section gives the derivations and implementation details omitted from Sec.~\ref{sec:method}. All coefficients and target distributions are constructed from detached model outputs. Backpropagation is performed only through the live student distribution $p_{\boldsymbol{\theta}}^t$.

\subsection{Execution Order and Feedback Conditioning}
\label{app:sdpo_details}

At optimizer step $k$, the self-distillation process inspired by~\citet{hubotter2026reinforcement} performs the following operations:

\begin{enumerate}
\item Sample a prompt $\mathbf{x}$ and generate one response
$\mathbf{y}\sim\pi_{\boldsymbol{\theta}_k}(\cdot\mid\mathbf{x})$
using the current student.
\item Construct privileged feedback
$\mathbf{c}=\mathcal{F}(\mathbf{x},\mathbf{y})$ for that response.
\item Score the fixed response with the live student conditioned on
$(\mathbf{x},\mathbf{y}_{<t})$ and with the EMA teacher conditioned on
$(\mathbf{x},\mathbf{c},\mathbf{y}_{<t})$.
\item Backpropagate the reverse-KL loss and update the student.
\item Update the EMA teacher from the newly updated student.
\end{enumerate}

The teacher therefore supplies feedback-conditioned supervision on samples visited by the student. It neither produces the training trajectory nor receives gradients. FIRE preserves this rollout, student-update, and EMA-update ordering, while feedback construction and EMA-teacher scoring are performed only for incorrect responses after outcome routing.

\subsection{Exact Reverse-KL Logit Gradient}
\label{app:reverse_kl_gradient}
Let $p^t_{\boldsymbol{\theta}}=\operatorname{softmax}(\mathbf{z}^t)$ and let $q^t$ be a
detached target. The token-level reverse KL is
\begin{equation}
\mathrm{KL}\!\left(p^t_{\boldsymbol{\theta}}\,\middle\|\,q^t\right)
=
\sum_v
p^t_{\boldsymbol{\theta}}(v)
\left[
\log p^t_{\boldsymbol{\theta}}(v)-\log q^t(v)
\right].
\label{eq:app_reverse_kl}
\end{equation}
Using
\begin{equation}
\frac{\partial p^t_{\boldsymbol{\theta}}(v)}{\partial z^t(u)}
=
p^t_{\boldsymbol{\theta}}(v)
\bigl(
\mathbf{1}[u=v]-p^t_{\boldsymbol{\theta}}(u)
\bigr),
\end{equation}
the logit gradient evaluated at the detached student snapshot
$p^t(v)=\operatorname{sg}[p^t_{\boldsymbol{\theta}_k}(v)]$ is
\begin{equation}
\bigl[\mathbf{g}^t(q^t)\bigr]_v
=
p^t(v)
\left[
\log\frac{p^t(v)}{q^t(v)}
-
\mathbb{E}_{u\sim p^t}\!\left[
\log\frac{p^t(u)}{q^t(u)}
\right]
\right].
\label{eq:app_gradient}
\end{equation}
The expectation centers the gradient so that
$\sum_v \bigl[\mathbf{g}^t(q^t)\bigr]_v=0$. FIRE compares feedback fields using this gradient
because it represents the update the target would actually apply to the current
student. Probability changes in regions assigned negligible student mass have little
immediate effect, whereas smaller changes on high-mass tokens may strongly affect
training.

\subsection{Fisher Scale and Learning-Rate Calibration}
\label{app:fisher_derivation}

For a categorical student distribution $p^t$, the softmax Fisher matrix is
\begin{equation}
\mathbf{F}(p^t)
=
\operatorname{diag}(p^t)-p^t(p^t)^\top,
\end{equation}
whose trace is
\begin{equation}
\phi^t
=
\operatorname{tr}\mathbf{F}(p^t)
=
1-\lVert p^t\rVert_2^2.
\label{eq:app_fisher}
\end{equation}
For an on-policy token $y_t\sim p^t$, the hard-label logit gradient is
$p^t-\mathbf{e}_{y_t}$, and, using
$\mathbb{E}_{y_t\sim p^t}[p^t(y_t)]=\lVert p^t\rVert_2^2$,
\begin{align}
\mathbb{E}_{y_t\sim p^t}
\lVert p^t-\mathbf{e}_{y_t}\rVert_2^2
&=
\mathbb{E}_{y_t\sim p^t}
\left[
1+\lVert p^t\rVert_2^2-2p^t(y_t)
\right]
\nonumber\\
&=
1+\lVert p^t\rVert_2^2-2\lVert p^t\rVert_2^2
=
1-\lVert p^t\rVert_2^2
=
\phi^t.
\label{eq:app_fisher_identity}
\end{align}
Therefore, $\sqrt{\phi^t}$ is the root-mean-square on-policy hard-label gradient norm. FIRE defines
\begin{equation}
s_k
=
\max\left(1,\frac{a_k}{a_0}\right),
\qquad
\rho_k^t
=
\frac{\sqrt{\phi^t}}{s_k}.
\label{eq:app_radius}
\end{equation}
When the scheduled learning rate is at or below the hyperparameter nominal rate value, FIRE retains the Fisher scale. When $a_k>a_0$, the radius contracts inversely with the learning-rate increase. The maximum prevents a decaying scheduler from enlarging the permitted gradient.

\subsection{Correct-Branch Re-weighting as Gradient Projection}
\label{app:correct_projection}

For a verified token, define
\begin{equation}
\mathbf{g}_+^t
=
p^t-\mathbf{e}_{y_t},
\qquad
\beta^t
=
\min\!\left(
1,
\frac{\rho_k^t}{\lVert\mathbf{g}_+^t\rVert_2}
\right).
\end{equation}
Because $\beta^t$ is detached from the computation graph,
\begin{equation}
\nabla_{\mathbf{z}^t}\ell_+^t
=
\beta^t
\bigl(
p^t-\mathbf{e}_{y_t}
\bigr).
\end{equation}
Consequently,
\begin{equation}
\left\lVert
\nabla_{\mathbf{z}^t}\ell_+^t
\right\rVert_2
=
\min\!\left(
\lVert\mathbf{g}_+^t\rVert_2,
\rho_k^t
\right).
\label{eq:app_correct_bound}
\end{equation}
Below the radius, FIRE is identical to ordinary on-policy SFT. Above it, the gradient lands exactly on the boundary without changing its direction.

We note the exact activation condition for the correct branch. Since
$\lVert\mathbf{g}_+^t\rVert_2^2=1+\lVert p^t\rVert_2^2-2p^t(y_t)$, clipping
at the hyperparameter ($s_k=1$, $(\rho_k^t)^2=\phi^t$) occurs
precisely when
\begin{equation}
p^t(y_t)<\lVert p^t\rVert_2^2
=
\mathbb{E}_{u\sim p^t}\left[p^t(u)\right].
\end{equation}
Clipping is therefore applied exactly to verified tokens that the current
student assigned below-average probability, i.e.\ to the tokens carrying the
largest hard-label gradient in an already correct trace. This is intended:
the radius equalizes the per-token step scale between the two branches, so
that a single surprising token inside a correct response cannot dominate the
update relative to the feedback-conditioned branch. The cost is that the
most informative correct tokens are the ones rescaled, which trades
per-token learning speed on the correct branch for comparability of step
size across branches.

\subsection{Exact Leave-One-Field-Out Influence}
\label{app:field_attribution}

For field $b_j$, define the full-versus-removed log-probability residual
\begin{equation}
r_j^t(v)
=
\log q_{-j}^t(v)
-
\log q_{\mathbf{c}}^t(v).
\end{equation}
Subtracting \eqref{eq:app_gradient} for the two teacher views gives
\begin{equation}
\bigl[
\mathbf{g}_{\mathbf{c}}^t
-
\mathbf{g}_{-j}^t
\bigr](v)
=
p^t(v)
\left[
r_j^t(v)
-
\mathbb{E}_{u\sim p^t} \left[ r_j^t(u) \right]
\right].
\label{eq:app_field_contrast}
\end{equation}
Therefore,
\begin{equation}
E_j^t
=
\left\lVert
\mathbf{g}_{\mathbf{c}}^t
-
\mathbf{g}_{-j}^t
\right\rVert_2^2
\end{equation}
is the exact change in the local training signal caused by removing field $b_j$. FIRE multiplies this influence by
\begin{equation}
\chi^t
=
\begin{cases}
\displaystyle
\frac{
    \left[E_{\mathbf{c}}^t-(\rho_k^t)^2\right]_+
}{
    E_{\mathbf{c}}^t
},
& E_{\mathbf{c}}^t>0,
\\[8pt]
0,
& E_{\mathbf{c}}^t=0,
\end{cases}
\end{equation}
where $E_{\mathbf{c}}^t=\lVert\mathbf{g}_{\mathbf{c}}^t\rVert_2^2$. This agrees with \eqref{eq:fire_evidence} in every case, and $\chi^t\in[0,1)$ since $(\rho_k^t)^2>0$ whenever $p^t$ is non-deterministic. The numerator is exactly the full teacher's gradient energy outside the permitted energy ball. If the complete feedback-conditioned target is already inside the ball, all field weights vanish and the full teacher is retained.

With
\begin{equation}
Z^t
=
(\rho_k^t)^2+\sum_jw_j^t,
\qquad
\alpha_0^t
=
\frac{(\rho_k^t)^2}{Z^t},
\qquad
\alpha_j^t
=
\frac{w_j^t}{Z^t},
\end{equation}
adopting the convention $\alpha_0^t=1$, $\alpha_j^t=0$ in the degenerate case $Z^t=0$, the attribution target is the normalized geometric mean
\begin{equation}
q_A^t(v)
=
\frac{
q_{\mathbf{c}}^t(v)^{\alpha_0^t}
\prod_{j=1}^{B}q_{-j}^t(v)^{\alpha_j^t}
}{
\sum_{u\in\mathcal{V}}
q_{\mathbf{c}}^t(u)^{\alpha_0^t}
\prod_{j=1}^{B}q_{-j}^t(u)^{\alpha_j^t}
}.
\label{eq:app_attribution_target}
\end{equation}
Its log probability satisfies
\begin{equation}
\log q_A^t(v)
=
C^t
+
\alpha_0^t\log q_{\mathbf{c}}^t(v)
+
\sum_j\alpha_j^t\log q_{-j}^t(v),
\end{equation}
where $C^t$ is constant across vocabulary elements. Since \eqref{eq:app_gradient} is affine in $\log q$ after centering, and since $\alpha_0^t+\sum_j\alpha_j^t=1$,
\begin{equation}
\mathbf{g}^t(q_A^t)
=
\alpha_0^t\mathbf{g}_{\mathbf{c}}^t
+
\sum_j\alpha_j^t\mathbf{g}_{-j}^t.
\label{eq:app_gradient_average}
\end{equation}
The geometric mean therefore converts the field-attribution coefficients into an exact weighted average of the corresponding training signals. An arithmetic probability mixture would not preserve this identity.

Two consequences of \eqref{eq:app_gradient_average} arise. Firstly, writing $\boldsymbol{\delta}_j^t=\mathbf{g}_{\mathbf{c}}^t-\mathbf{g}_{-j}^t$, the attribution step can be expressed as
\begin{equation}
\mathbf{g}^t(q_A^t)
=
\mathbf{g}_{\mathbf{c}}^t
-
\frac{\chi^t}{Z^t}
\sum_j
\lVert\boldsymbol{\delta}_j^t\rVert_2^2\,
\boldsymbol{\delta}_j^t. 
\end{equation}
Second, because \eqref{eq:app_gradient_average} is a convex combination,
\begin{equation}
\lVert\mathbf{g}^t(q_A^t)\rVert_2
\leq
\max\left(
\lVert\mathbf{g}_{\mathbf{c}}^t\rVert_2,
\max_j\lVert\mathbf{g}_{-j}^t\rVert_2
\right),
\end{equation}
but $\lVert\mathbf{g}^t(q_A^t)\rVert_2>\lVert\mathbf{g}_{\mathbf{c}}^t\rVert_2$ is possible when some $\boldsymbol{\delta}_j^t$ is anti-aligned with $\mathbf{g}_{\mathbf{c}}^t$. Attribution is therefore a reweighting of the correction direction and carries no norm guarantee on its own. The radius is enforced by the subsequent projection.

\subsection{Incorrect-Branch Radial Projection}
\label{app:failure_projection}

Let
\begin{equation}
\mathbf{g}_A^t
=
\mathbf{g}^t(q_A^t),
\qquad
\eta^t
=
\min\left(
1,
\frac{\rho_k^t}{\lVert\mathbf{g}_A^t\rVert_2}
\right),
\end{equation}
with $\eta^t=1$ for a zero gradient. FIRE constructs
\begin{equation}
q_{\mathrm{FIRE}}^t(v)
=
\frac{
p^t(v)^{1-\eta^t}
q_A^t(v)^{\eta^t}
}{
\sum_{u\in\mathcal{V}}
p^t(u)^{1-\eta^t}
q_A^t(u)^{\eta^t}
}.
\label{eq:app_fire_target}
\end{equation}
Up to a vocabulary-independent normalizer $C_\eta^t$,
\begin{equation}
\log q_{\mathrm{FIRE}}^t
=
(1-\eta^t)\log p^t
+
\eta^t\log q_A^t
+
C_\eta^t ,
\end{equation}
so the log-ratio appearing in \eqref{eq:app_gradient} reduces to
\begin{equation}
\log p^t-\log q_{\mathrm{FIRE}}^t
=
\eta^t\left(\log p^t-\log q_A^t\right)
-
C_\eta^t .
\end{equation}
The constant $C_\eta^t$ cancels under the centering in \eqref{eq:app_gradient}, and the remaining factor $\eta^t$ passes through linearly, yielding
\begin{equation}
\mathbf{g}^t(q_{\mathrm{FIRE}}^t)
=
\eta^t\mathbf{g}_A^t.
\end{equation}
Therefore,
\begin{equation}
\left\lVert
\mathbf{g}^t(q_{\mathrm{FIRE}}^t)
\right\rVert_2
=
\min\!\left(
\lVert\mathbf{g}_A^t\rVert_2,
\rho_k^t
\right).
\label{eq:app_failure_bound}
\end{equation}
Attribution changes which feedback-dependent direction is followed, and radial projection then restricts how strongly that direction is applied. The two stages are sequential rather than independent, since $\eta^t$ is determined by $\lVert\mathbf{g}_A^t\rVert_2$ and therefore by the outcome of the attribution step.

\subsection{Unified Bound and Its Scope}
\label{app:unified_bound}

Equations~\ref{eq:app_correct_bound} and~\ref{eq:app_failure_bound} imply that either routed branch satisfies
\begin{equation}
\lVert\mathbf{g}_{\mathrm{branch}}^t\rVert_2
\leq
\rho_k^t
=
\frac{
\sqrt{\phi^t}
}{
\max(1,a_k/a_0)
}.
\end{equation}
When $a_k\geq a_0$,
\begin{equation}
a_k
\lVert\mathbf{g}_{\mathrm{branch}}^t\rVert_2
\leq
a_0\sqrt{\phi^t}.
\label{eq:app_step_bound}
\end{equation}
This is a local output-logit bound. It does not directly bound the full parameter update, next-token KL, or global convergence, which additionally depend on the model Jacobian, optimizer preconditioning and momentum, gradient accumulation, and parameter-gradient clipping.

\subsection{Computation and Memory}
\label{app:implementation}

Correct responses require one student rollout and one student scoring pass, but no teacher evaluation. Incorrect responses use the same rollout, one live student scoring pass, one full-feedback EMA-teacher score, and $B$ leave-one-field-out teacher scores. None of the teacher passes generates a new response.

The leave-one-field-out views can be streamed. Define
\begin{equation}
S_w^t=\sum_jw_j^t,
\qquad
R_w^t
=
\sum_jw_j^t
\left(
\log q_{-j}^t-\log q_{\mathbf{c}}^t
\right).
\end{equation}
Then, since $1-S_w^t/Z^t=(\rho_k^t)^2/Z^t$, the attribution target can be computed as
\begin{equation}
\log q_A^t
=
\log\operatorname{softmax}\left(
\log q_{\mathbf{c}}^t
+
\frac{
R_w^t
}{
(\rho_k^t)^2+S_w^t
}
\right).
\end{equation}
The implementation therefore stores one residual accumulator rather than all $B$ removed-field distributions simultaneously. FIRE introduces no additional student rollout, student forward pass, or backward pass beyond the underlying self-distillation loop.

Two numerical caveats apply. First, the exact gradient identities assume full-vocabulary reverse KL, while truncating the vocabulary support makes them approximate. Second, because \eqref{eq:app_attribution_target} is a geometric mean, $\log q_A^t(v)$ diverges whenever any teacher view with $\alpha_j^t>0$ assigns vanishing mass to $v$, so $q_A^t$ effectively inherits the narrowest support among the active views. In practice all teacher log-probabilities are floored at a small constant before accumulation into $R_w^t$, which bounds the residual and leaves the identities in Appendices~\ref{app:field_attribution} and~\ref{app:failure_projection} accurate to that floor.

\end{document}